\documentclass{article}

\usepackage{hyperref}
\usepackage{url}
\usepackage{amsmath}
\usepackage{amssymb}
\usepackage{amsfonts}
\usepackage{multirow}
\usepackage{multicol}
\usepackage[utf8]{inputenc}
\usepackage{graphicx}
\usepackage{adjustbox}
\usepackage{float}
\usepackage{pgfplots}
\usepackage{subcaption}
\usepackage{wrapfig}
\usepackage{bm}

\usepackage{amsthm}

\newtheorem{prop}{Proposition}
\newtheorem{remark}{Remark}
\newtheorem{theorem}{Theorem}
\newtheorem{Def}{Definition}

\usepackage{microtype}
\usepackage{graphicx}
\usepackage{booktabs} 

\usepackage{hyperref}

\newcommand{\eg}{\emph{e.g. }}
\newcommand{\ie}{\emph{i.e. }}

\newcommand{\Diag}{\mathrm{Diag}}
\newcommand{\diag}{\mathrm{diag}}
\newcommand{\IN}{\mathrm{IN}}
\newcommand{\BN}{\mathrm{BN}}

\newcommand{\Expe}{\mathbb{E}}
\newcommand{\Dvar}{\mathrm{D}}

\newcommand*{\tran}{{^{\mkern-1.5mu\mathsf{T}}}}
\newcommand{\R}{\mathbb{R}}
\newcommand{\pp}{\bm{p}}

\usepackage[accepted]{icml2020}

\newcommand{\equref}[1]{Eqn.(\ref{#1})}

\begin{document}

\twocolumn[
\icmltitle{Channel Equilibrium Networks for Learning Deep Representation}



\icmlsetsymbol{equal}{*}

\begin{icmlauthorlist}
\icmlauthor{Wenqi Shao}{equal,to}
\icmlauthor{Shitao Tang}{equal,goo}
\icmlauthor{Xingang Pan}{to}
\icmlauthor{Ping Tan}{goo}
\icmlauthor{Xiaogang Wang}{to}
\icmlauthor{Ping Luo}{ed}
\end{icmlauthorlist}

\icmlaffiliation{to}{The Chinese University of Hong Kong}
\icmlaffiliation{goo}{Simon Fraser University}
\icmlaffiliation{ed}{The University of Hong Kong}

\icmlcorrespondingauthor{Wenqi Shao}{weqish@link.cuhk.edu.hk}
\icmlcorrespondingauthor{Ping Luo}{pluo@cs.hku.hk}


\vskip 0.3in
]



\printAffiliationsAndNotice{\icmlEqualContribution} 

\begin{abstract}
Convolutional Neural Networks (CNNs) are typically constructed by stacking multiple building blocks, each of which contains a normalization layer such as batch normalization (BN) and a rectified linear function such as ReLU. 
However, this work shows that the combination of normalization and rectified linear function leads to inhibited channels, which have small magnitude and contribute little to the learned feature representation, impeding the generalization ability of CNNs.
Unlike prior arts that simply removed the inhibited channels, we propose to ``wake them up'' during training by designing a novel neural building block, termed Channel Equilibrium (CE) block, which enables channels at the same layer to contribute equally to the learned representation. We show that CE is able to prevent inhibited channels both empirically and theoretically.
CE has several appealing benefits. (1) It can be integrated into many advanced CNN architectures such as ResNet and MobileNet, outperforming their original networks. (2) CE has an interesting connection with the Nash Equilibrium, a well-known solution of a non-cooperative game. (3) Extensive experiments show that CE achieves state-of-the-art performance on various challenging benchmarks such as ImageNet and COCO.

\end{abstract}

\section{Introduction}
Normalization methods such as batch normalization (BN) \citep{ioffe2015batch}, layer normalization (LN) \citep{ba2016layer} and instance normalization (IN) \citep{ulyanov2016instance} are important components for a wide range of tasks such as image classification \citep{ioffe2015batch}, object detection \citep{he2017mask}, and image generation \citep{miyato2018spectral}. 
They are often combined with rectified linear activation functions such as rectified linear unit (ReLU) \citep{glorot2011deep,nair2010rectified}, exponential linear unit (ELU) \citep{clevert2015fast} and leaky ReLU (LReLU) \citep{maas2013rectifier} and used in many recent advanced convolutional neural networks (CNNs).
The combination of normalization and rectified unit becomes one of the most popular building block for CNNs. 
\begin{figure*}
\begin{subfigure}{.33\textwidth}
\centering
\includegraphics[width=0.75\textwidth]{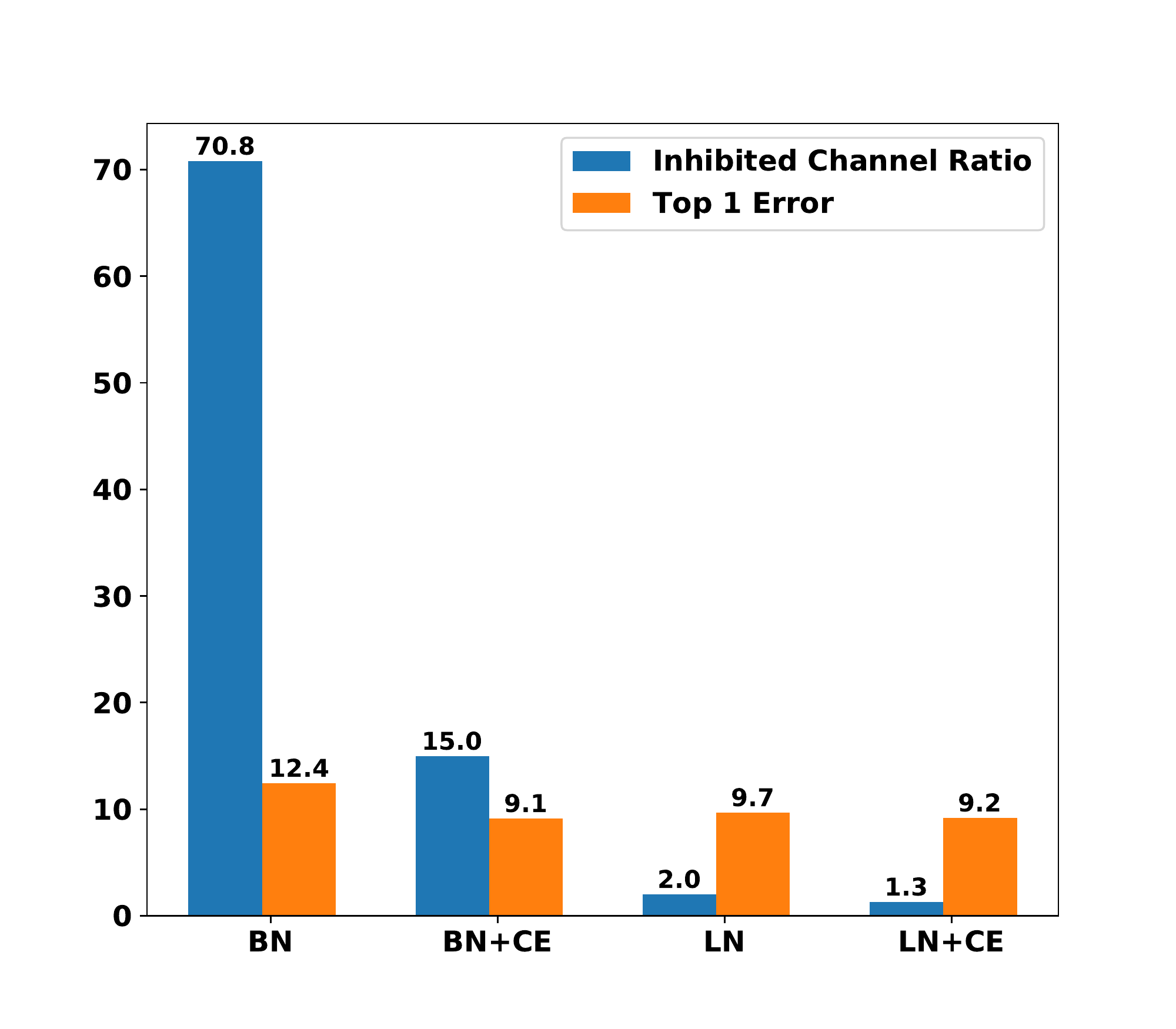}
\caption{CE improves BN and LN.}
\end{subfigure}
\begin{subfigure}{.33\textwidth}
\centering
\includegraphics[width=0.75\textwidth]{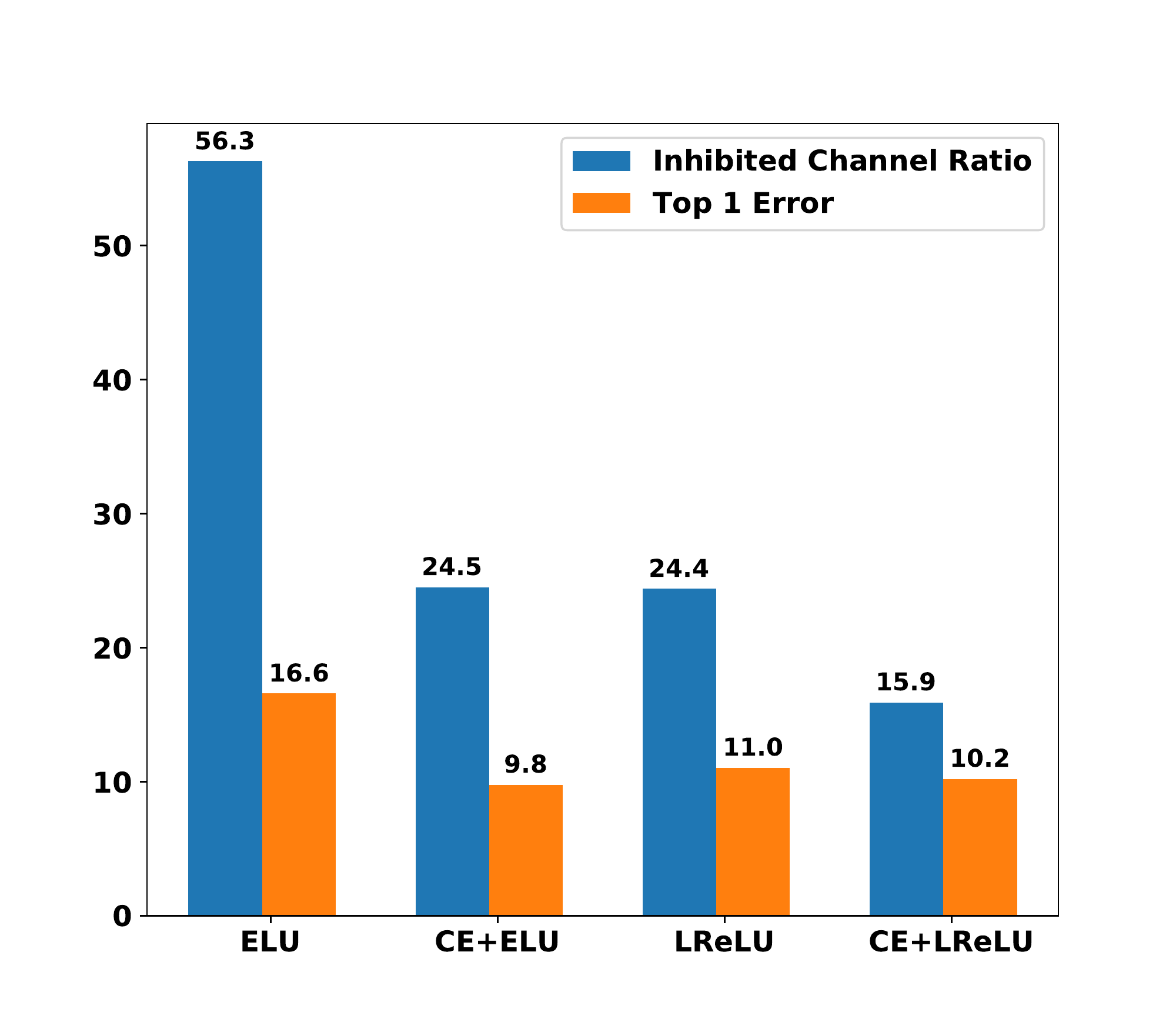}
\caption{CE improves ELU and LReLU.}
\end{subfigure}
\begin{subfigure}{.33\textwidth}
\centering
\includegraphics[width=0.75\textwidth]{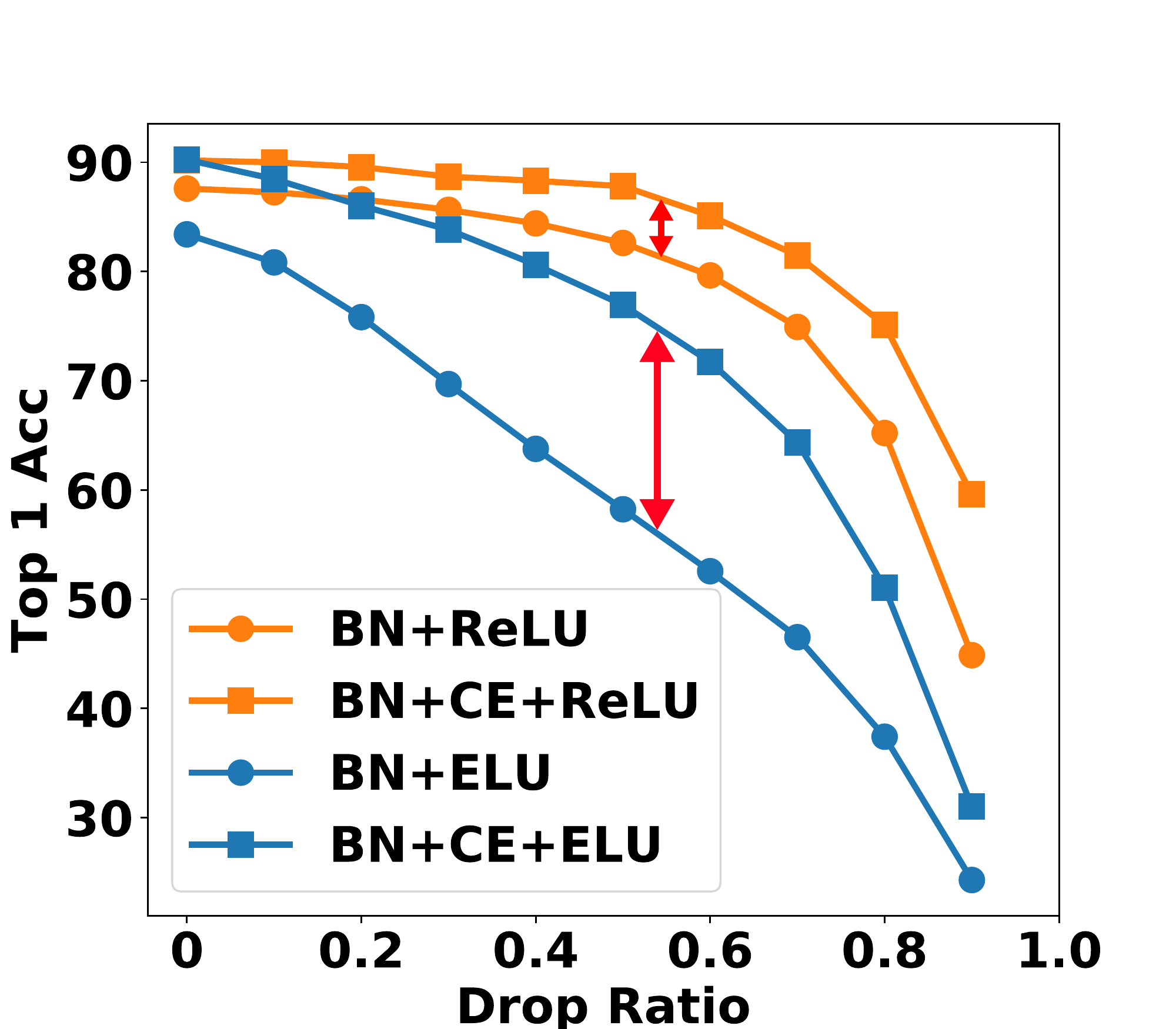}
\caption{Cumulative Ablation Curves}
\end{subfigure}
\vspace{-0.3cm}
\caption{CE can improve many different normalization methods and rectified linear activation functions. For example, in (a-c), CE is used to train VGGNet  \citep{simonyan2014very} on CIFAR10 \citep{krizhevsky2009learning} with different normalizers and rectified units. (a) shows that the numbers of inhibited channels (\ie channel features with values $<10^{-2}$) are greatly reduced by applying CE with BN and LN, whose top-1 test errors are also reduced by using CE. In (a), ReLU is the activation.
(b) shows similar phenomena where CE decreases inhibited channels and test errors compared to the ordinary ELU and LReLU functions. In (b), BN is the normalization method.
(c) demonstrates that CE can encourage channels of BN+ReLU or ELU to contribute more equally to the network's prediction, by using cumulative ablation curve \citep{morcos2018importance}, where accuracies are evaluated by randomly ablating channels (\ie set their values to zeros) with an increasing ratio from `$0$' to `$1$'. When the ratio approaches `0.9', most channels are set to zeroed values, resulting in the worst accuracy. We see that CE presents a more gentle accuracy drop curve, implying that it makes the network less reliant on specific channels.
}
\label{fig1:sparse-top1first}
\vspace{-0.4cm}
\end{figure*}

However, recent studies showed that the above building block leads to inhibited channels (as known as ``channel collapse'') after training a CNN, where a significant amount of feature channels always produce small values \citep{mehta2019implicit} as shown in Fig.\ref{fig1:sparse-top1first}(a\&{b}). These inhibited channels contribute little to the learned feature representation, making the network more reliant on the remaining channels, which impedes its generalization ability as shown in \citep{morcos2018importance}.
For example, the lottery hypothesis \citep{frankle2018lottery} found that when a CNN is over-parameterized, it always contains unimportant (``dead'') channels whose feature values are extremely small. 
Although these inhibited channels could be pruned in training to reduce the model size, it would lead to the limited generalization ability of the network \citep{yu2018slimmable,he2017channel}.

Instead of simply removing the inhibited ``dead'' channels, this work investigates an alternative to ``wake them up'' by proposing a novel neural building block, termed Channel Equilibrium (CE), to replace the ordinary combination of normalization and rectified units. 
CE encourages channels at the same layer of a network to contribute more equally in representation learning.
With CE, all channels are useful in the learned representation, preventing CNNs from relying on specific channels and thus enhancing the generalization ability. 
For example, Fig.\ref{fig1:sparse-top1first} shows that CE not only reduces the number of inhibited channels but encourages all channels to contribute equally to network's prediction, when different combinations of normalization approaches and rectified units are presented, consistently improving their generalization to testing samples.

The main \textbf{contributions} of this work are three-fold. (1) We propose a novel neural building block for CNNs, termed Channel Equilibrium (CE), which encourages all channels to contribute equally to the learned feature representation. In theory, 
we show an interesting connection between CE and Nash Equilibrium, which is a well-known solution in game theory.
%
%
%
%
%
(2) CE can significantly improve the generalization of existing networks with merely small computational overhead by plugging it into various advanced CNN architectures. For example, when CE is integrated into ResNet50 \citep{he2016deep} and MobileNetv2 \citep{sandler2018mobilenetv2}, 
the resulting networks substantially outperform the original networks by $1.7$\% and $2.1$\% top-1 accuracy on ImageNet \citep{russakovsky2015imagenet}, while merely introducing small extra computation. Specifically, the improvement of ResNet50+CE over ResNet50+BN is 70\% larger than that of ResNet50+Squeeze-and-Excitation block \citep{hu2018squeeze} (\ie 1.7\% versus 1.0\%).
(3)
The learned representation of CE can be well generalized to many other tasks such as object detection and segmentation. For example, CE trained with Mask RCNN \citep{he2017mask} using ResNet50 as backbone improves the AP metric on the MS-COCO dataset \citep{lin2014microsoft} by $3.4$ compared to its counterpart.

\section{Notation and Preliminary}

This section presents the notations and backgrounds of normalization methods and rectified units.

\textbf{Notations.} We use regular letters to denote scalars such as `$x$', and use bold letters to denote vectors (\eg vector, matrix, and tensor) such as `$\bm{x}$'. For CNNs, we employ a 4D tensor, $\bm{x}\in\R^{N\times C\times H \times W}$, to represent the feature map in a layer, where $N, C, H$ and $W$ indicate sample size, channel size, height and width of a channel respectively.
For example, $x_{ncij}$ denotes a pixel at location $(i,j)$ in the $c$-th channel of the $n$-th sample. 

\textbf{Overview.} The recently advanced building block of CNNs consists of a normalization layer and a rectified linear function denoted as $g(\cdot)$. We have
\begin{equation}\label{eq:bn+relu0}
\begin{split}
y_{ncij}&=g(\tilde{x}_{ncij}), \,\text{where}\\
\tilde{x}_{ncij}&=\gamma_c\bar{x}_{ncij}+\beta_c,\quad \bar{x}_{ncij}=(x_{ncij}-\mu_{k})/\sigma_{k}.
\end{split}
\end{equation}
In Eqn.(\ref{eq:bn+relu0}), $y_{ncij}$ denotes the output value after applying rectified activation function and normalization method. 
$k\in \Omega = \{\IN, \BN, \cdots\}$ where $\Omega$ indicates a set of normalization methods. $\mu_{k}$ and $\sigma_{k}$ are mean and standard deviation estimated by using the normalizer $k$. Moreover, $\tilde{x}_{ncij}$ and $\bar{x}_{ncij}$ respectively represent the features after normalization and standardization (\ie with zeroed mean and unit standard deviation). For each channel, ${\gamma_c}$ and ${\beta_c}$ are two parameters, which re-scale and re-shift the standardized features $\bar{x}_{ncij}$. Furthermore, $g(\cdot)$ denotes a rectified linear function. For instance, we have $g(x)=x\cdot\mathbf{1}_{x\geq 0}+ax\cdot\mathbf{1}_{x<0}$. It represents ReLU \citep{nair2010rectified} when $a=0$, while it represents leaky ReLU (LReLU) \citep{maas2013rectifier} when $a\in (0,1)$.

\textbf{Inhibited Channels.} \equref{eq:bn+relu0} shows that many normalization approaches perform an affine transformation by using the parameters ${\gamma_c}$ and ${\beta_c}$ for each channel. Previous work \cite{mehta2019implicit} shows that after training,  amounts of $\gamma_c$ and $y_{ncij}$ for all $i\in [H]$ and $j\in [W]$ would get small. We see this by treating $\bar{x}_{ncij}$ in Eqn.(\ref{eq:bn+relu0}) as a standard Gaussian random variable following \citep{arpit2016normalization}. When the value of $\gamma_c$ becomes small, Remark \ref{remark:1} tells us that the mean and the variance of the channel output $y_{ncij}$ would also be small (proof is provided in Appendix Sec.\ref{sec:appendixremark1}). In this case, the $c$-th channel becomes inhibited and contributes little to the representation learning.
For evaluation, this paper treats those channels with magnitudes smaller than $10^{-2}$ as inhibited channels. We observe that inhibited channels largely emerge in many different combinations of normalizations and rectified units, including BN \citep{ioffe2015batch}, IN \citep{ulyanov2016instance}, LN \citep{ba2016layer}, ReLU, ELU \citep{clevert2015fast} and LReLU \citep{maas2013rectifier} as shown in Fig.\ref{fig1:sparse-top1first}(a{\&}b). The existence of inhibited channels makes the network rely more on the remaining activated channels, impeding the generalization of CNNs \citep{morcos2018importance}.

\vspace{-3pt}
\begin{remark}\label{remark:1}
Let a random variable $z\sim\mathcal{N}(0,1)$ and $y=max\{0,\gamma_cz+\beta_c\}$. Then we have  $\mathbb{E}_z[y]=0$ and $ \mathbb{E}_z[y^2]=0$ if and only if  $\beta_c\leq 0$ and $\gamma_c$ sufficiently approaches $0$.
\end{remark}
\vspace{-3pt}
\textbf{Decorrelation.}
Although the above ELU and LReLU extend the ReLU activation function by making its negative part has a non-zero slope, they are not able to prevent inhibited channels. Different from these methods, this work prevents inhibited channels by decorrelation operation performed after the normalization layer. Typically, a decorrelation operator is expressed as the inverse square root of the covariance matrix, denoted as $\bm{\Sigma}^{-\frac{1}{2}}$ where $\bm{\Sigma}$ is the covariance matrix and is usually estimated over a minibatch of samples \cite{huang2018decorrelated,huang2019iterative}. This work discovers that decorrelating feature channels after normalization layer 
can increase the magnitude of all the feature channels, making all channels useful in the learned representation. 

Furthermore, suppose that every single channel aims to contribute to the learned feature representation, we show that decorrelating feature channels after the normalization method can be connected with the Nash Equilibrium for each instance.
In this sense, constructing a decorrelation operator for every single sample is also crucial for representation learning  \citep{yang2019cross}.
As presented in the below section, the proposed Channel Equilibrium (CE) module is carefully designed by exploring a dynamic decorrelation operator conditioned on each instance sample.

\section{Channel Equilibrium (CE) Block}

This section introduces the CE block, which contains a branch of batch decorrelation (BD) and a branch of instance reweighting (IR). 
We show that the CE block can increase the magnitude of feature channels. We also show the connection between the CE block and the Nash Equilibrium.

In particular, a CE block is a computational unit that encourages all channels to contribute to the feature representation by decorrelating feature channels. 
Unlike previous methods \citep{huang2018decorrelated,huang2019iterative} that decorrelated features after the convolutional layer given a minibatch of samples, CE conditionally decorrelates features after the normalization layer for each sample. Rewriting Eqn.(\ref{eq:bn+relu0}) into a vector, we have the formulation of CE
\vspace{-3pt}
\begin{equation}\label{eq:CE}
\pp_{nij}=\bm{D}_n^{-\frac{1}{2}}(\Diag(\bm{\gamma})\bar{\bm{x}}_{nij}+\bm{\beta})
\end{equation} 
where $\pp_{nij}\in\R^{ C\times1}$ is a vector of $C$ elements that denote the output of CE for the $n$-th sample at location $(i,j)$ for all channels. $\bm{D}_n^{-\frac{1}{2}}$ is a decorrelation operator and $\bm{D}_n$ is the covariance matrix defined in CE. The subscript $n$ is the sample index, suggesting that the decorrelation operator is performed for each sample but not a minibatch of samples. 
In Eqn.\eqref{eq:CE}, $\bm{\bar{x}}_{nij}\in \R^{C\times1}$ is a vector by stacking elements from all channels of $\bar{x}_{ncij}$ into a column vector. $\bm{\gamma}\in \R^{C\times1}$ and $\bm{\beta} \in \R^{C\times1}$ are two vectors by stacking $\gamma_c$ and $\beta_c$ of all the channels respectively. $\Diag (\bm{\gamma})$ returns a diagonal matrix by using $\bm{\gamma}$ as diagonal elements.

To decorrelate the feature channels conditioned on each input, statistics of the channel dependency with respect to both the minibatch and each sample are embedded in the matrix $\bm{D}_n$. We achieve this by incorporating a covariance matrix $\bm{\Sigma}$ with an instance variance matrix, $\Diag(\bm{v}_n)$, where $\bm{v}_n\in\R^{C\times1}$ denotes the adaptive instance variances for all channels. In this way, we have
\vspace{-3pt}
\begin{equation}\label{eq:D_n}
\bm{D}_n=\lambda \bm{\Sigma}+(1-\lambda)\mathrm{Diag}(\bm{v}_n),\quad\,\,\bm{v}_n=f(\tilde{\bm{\sigma}}_n^2),
\end{equation}
where $\bm{\Sigma}\in\R^{C\times C}$ is estimated by a minibatch of samples after normalization,$\{\tilde{\bm{x}}_n\}_{n=1}^N$,    $\tilde{\bm{\sigma}}_n^2\in\R^{C\times1}$ is a vector of variance of the $n$-th instance estimated by using $\tilde{\bm{x}}_n$ for all channels \citep{ulyanov2016instance}, $f: \R^{C\times1}\rightarrow\R^{C\times1}$ models channel dependencies and returns an adaptive instance variance. And $\lambda\in(0,1)$ is a learnable ratio used to switch between the batch and the instance statistics. 

Given Eqn.\eqref{eq:D_n}, the decorrelation operator $\bm{D}_n^{-\frac{1}{2}}$ can be relaxed by using the Jensen inequality for matrix functions \citep{pevcaric1996power}. We have 
\vspace{-3pt}
\begin{equation}\label{eq:D_n-relax}
\begin{split}
\bm{D}_n^{-\frac{1}{2}}&=\left[\lambda \bm{\Sigma}+(1-\lambda)\mathrm{Diag}(\bm{v}_n)\right]^{-\frac{1}{2}}\\
&\preceq \lambda \underbrace{\bm{\Sigma}^{-\frac{1}{2}}}_{\textsuperscript{batch decorrelation}}+(1-\lambda) \underbrace{\left[\mathrm{Diag}(\bm{v}_n)\right]^{-\frac{1}{2}}}_{\textsuperscript{instance reweighting}},
\end{split}
\end{equation}
where $\bm{A}\preceq \bm{B}$ indicates $\bm{B}-\bm{A}$ is semi-definite positive. The above relaxation is made because of 
two reasons. (1) \emph{Reduce Computational Complexity}. It allows less computational cost for each training step since the relaxed form only needs to calculate the inverse of square root $\bm{\Sigma}^{-\frac{1}{2}}$ once, meanwhile the other branch $\Diag (\bm{v}_n)^{-\frac{1}{2}}$ is easy to compute. (2) \emph{Accelerate Inference.} $\bm{\Sigma}^{-\frac{1}{2}}$ is a moving-average statistic in inference, which can be absorbed into the previous layer, thus enabling fast inference. 

\begin{figure}
    \centering
    \includegraphics[width=1.0\linewidth]{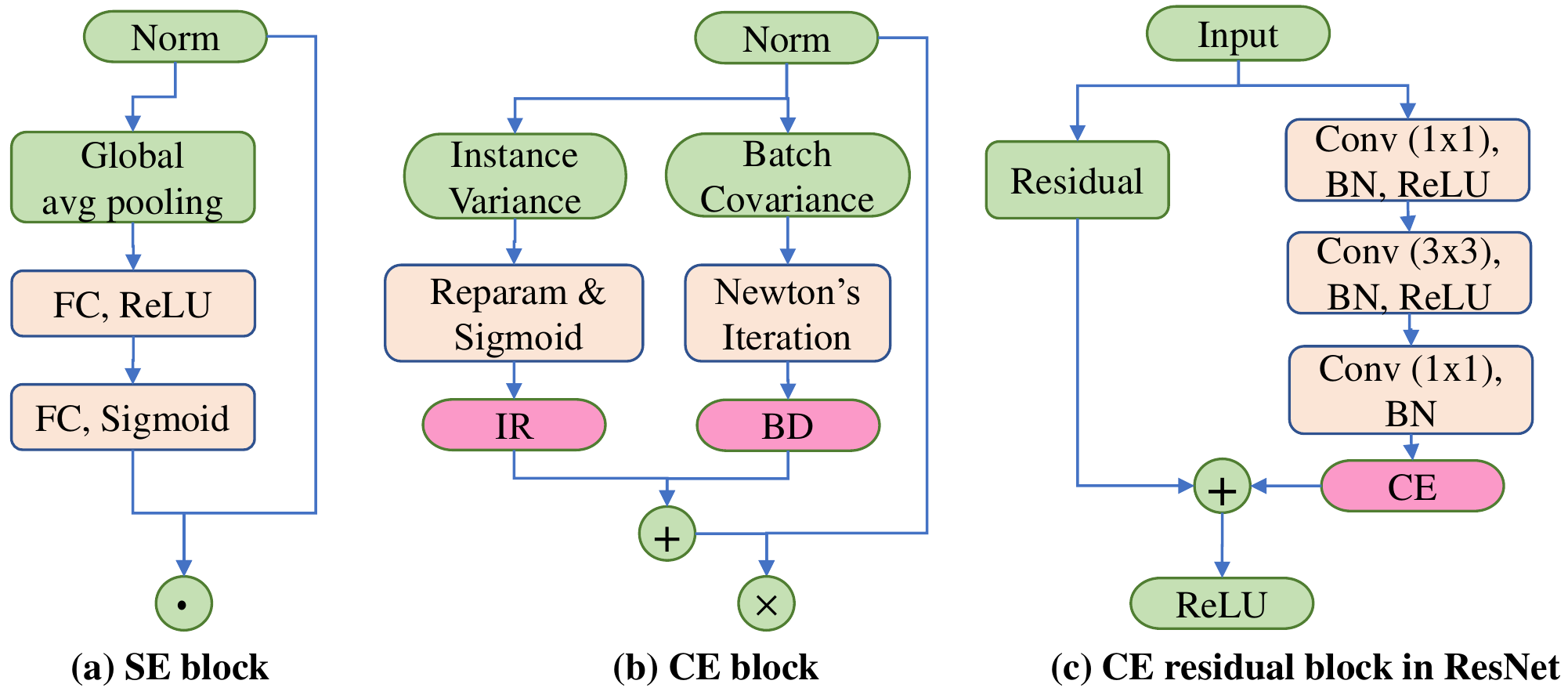}
    \vspace{-3pt}
    \caption{\textbf{Comparisons} of (a) SE block \citep{hu2018squeeze}, (b) CE block and (c) CE residual block in ResNet. $\odot$ denotes broadcast element-wise multiplication, $\textcircled{+}$ denotes broadcast elementwise addition and $\otimes$ denotes matrix multiplication. The SE block in (a) is not able to equalize feature representation, and it has larger computations and lower performance than (b). In (b), CE has two lightweight branches, BD and IR. (c) shows CE can be easily stacked into many advanced networks such as ResNet with merely small extra computation.}
    \label{fig:CE-SE}
\vspace{-10pt}
\end{figure}

In the following descriptions, we treat $\bm{\Sigma}^{-\frac{1}{2}}$ in Eqn.(\ref{eq:D_n-relax}) as batch decorrelation (BD) and treat $\left[\mathrm{Diag}(\bm{v}_n)\right]^{-\frac{1}{2}}$ as instance reweighting (IR). The former one performs decorrelation by using a covariance matrix estimated in an entire minibatch, 
while the latter one adjusts correlations among feature channels by reweighting each channel with the inverse square root of an adaptive variance for each instance. 
Integrating both of them yields a dynamic decorrelation operator conditioned on each instance in the CE bock whose forward representation is illustrated in Fig.\ref{fig:CE-SE}(b).

\vspace{-2pt}
\subsection{Batch Decorrelation (BD)}
Although many previous work \citep{huang2018decorrelated,huang2019iterative,pan2019switchable} have investigated decorrelation (whitening) methods by using the covariance matrix, all of them are applied in the normalization layer. Their drawback is that the channel features after whitening are still scaled by $\bm{\gamma}$ channel-wisely in the normalization layer, thus producing inhibited channels. Instead, CE is applied after the normalization layer (after $\bm{\gamma}$), which as will be shown, is able to explicitly prevent inhibited channels. We take batch normalization (BN) as an example to illustrate CE. Note that CE can be applied to any normalization methods and activation functions. 

Consider a tensor $\bar{\bm{x}}$ after a BN layer, it can be reshaped as $\bar{\bm{x}}\in\R^{C\times M}$ and $M=N\cdot H\cdot W$. Then the covariance matrix $\bm{\Sigma}$ of the normalized features $\tilde{\bm{x}}$ can be written as (details in Sec.\ref{sec:appendixA} of Appendix)
\vspace{-2pt}
\begin{equation}\label{eq:batch-co}
\bm{\Sigma}=\bm{\gamma}\bm{\gamma}\tran \odot \frac{1}{M}\bar{\bm{x}}\bar{\bm{x}}\tran,\vspace{-2pt}
\end{equation}
\vspace{-2pt}
where $\bar{\bm{x}}$ is a standardized feature  with zero mean and unit variance and $\odot$ indicates elementwise multiplication. 
It is observed that each element $\Sigma_{ij}$ represents the dependency between the $i$-th channel and the $j$-th channel, and it is scaled by $\gamma_i\gamma_j$ after normalization.

The BD branch requires computing $\bm{\Sigma}^{-\frac{
1}{2}}$, which usually uses eigen-decomposition or SVD, thus involving heavy computations \citep{huang2018decorrelated}. Instead, we adopt an efficient Newton's Iteration to obtain $\bm{\Sigma}^{-\frac{1}{2}}$ \citep{bini2005algorithms,higham1986newton}. Given a covariance matrix $\bm{\Sigma}$, Newton's Iteration calculates $\bm{\Sigma}^{-\frac{1}{2}}$ by following the iterations,
\vspace{-2pt}
\begin{equation}\label{eq:Newtoniter}
\left\{
\begin{array}{l}
\bm{\Sigma}_0=\bm{I}\\
\bm{\Sigma}_k=\frac{1}{2}(3\bm{\Sigma}_{k-1}-\bm{\Sigma}_{k-1}^3\bm{\Sigma}),\, k=1,2,\cdots,T.
\end{array}
\right.
\end{equation}
\vspace{-2pt}
where $k$ is the iteration index and $T$ is the iteration number ($T=3$ in our experiments). Note that the convergence of Eqn.(\ref{eq:Newtoniter}) is guaranteed if $\left\|\bm{I}-\bm{\Sigma}\right\|_2<1$ \citep{bini2005algorithms}. To satisfy this condition, $\bm{\Sigma}$ can be normalized by $\bm{\Sigma}/\mathrm{tr}(\bm{\Sigma})$ following  \citep{huang2019iterative}, where $\mathrm{tr}(\cdot)$ is the trace operator. In this way, the normalized covariance matrix can be written as $\bm{\Sigma}=\frac{\bm{\gamma}\bm{\gamma}\tran}{\left\|\bm{\gamma}\right\|_2^2} \odot \frac{1}{M}\bar{\bm{x}}\bar{\bm{x}}\tran$. To sum up, in the training stage, the BD branch firstly calculates a normalized covariance matrix and then applies Newton's Iteration to obtain its inverse square root, reducing computational cost compared to the SVD decomposition. In the testing stage, BD can be merged into the convolutional layers, which merely adds small extra computation. 

\vspace{-2pt}
\subsection{Instance Reweighting (IR)}
Other than the BD branch, the decorrelation operator is also desired for each sample \citep{yang2019cross}. we achieve this by incorporating the BD with a branch of instance reweighting (IR) as shown in Eqn.(\ref{eq:D_n-relax}).

Specifically, the input of IR is denoted as $\tilde{\bm{\sigma}}^2_{n}\in\R^{C\times 1}$, which  can be computed as below (details in Appendix Sec.\ref{sec:appendixA})
\vspace{-2pt}
\begin{equation}\label{eq:inputAII}
\tilde{\bm{\sigma}}^2_{n}=\diag(\bm{\gamma}\bm{\gamma}\tran)\odot\frac{(\bm{\sigma}^2_{\IN})_{n}}{\bm{\sigma}^2_{\BN}},
\end{equation}
where $\diag (\bm{\gamma}\bm{\gamma}\tran) \in \R^{C\times 1}$ extracts the diagonal of the given matrix,   $(\bm{\sigma}^2_{\IN})_{n}\in\R^{C\times 1}$ and $\bm{\sigma}^2_{\BN}\in\R^{C\times 1}$ represent the variances estimated by using IN \citep{ulyanov2016instance} and BN \citep{ioffe2015batch} respectively. In Eqn.(\ref{eq:inputAII}), the vector division is applied elementwisely.
Similar to Eqn.(\ref{eq:batch-co}), the input of IR is scaled by $\gamma_c^2$ for the $c$-th channel. 

The IR branch returns an inverse square root of an adaptive instance inverse, denoted as $\left[\mathrm{Diag}(\bm{v}_n)\right]^{-\frac{1}{2}}$, which is used to adjusts correlations among feature channels. It needs to satisfy two requirements. First, 
note that $\bm{v}_n=f(\tilde{\bm{\sigma}_n^2})$  from Eqn.(\ref{eq:D_n}), while $\tilde{\bm{\sigma}}_n^2$ is just  a vector of variances calculated within each channel. 
To adjust correlations by IR branch,
the dependencies among channels should be embedded in transformation $f$ for each sample.  Second, the output of IR should have the same magnitude as the inverse square root of variance or covariance in the BD branch such that neither of them is dominant in CE. 
 To achieve the above, a reparameterization trick is employed to generate the inverse square root of instance variance. Let $s=\frac{1}{NC}\sum_{n,c}^{N,C}(\tilde{\bm{\sigma}}_{n}^2)_c$ be the estimate of variance for all channels and all samples in a minibatch, the transformation $f$ in Eqn.(\ref{eq:D_n}) can be reparameterized as below,
 \vspace{-3pt}
\begin{equation}
\left[\mathrm{Diag}(\bm{v}_n)\right]^{-\frac{1}{2}}=\Diag(\mathrm{Sigmoid}(\tilde{\bm{\sigma}}_n^2;\bm{\theta}))\cdot s^{-\frac{1}{2}},\label{eq:reparam1}\\
\end{equation} 
where  $s^{-\frac{1}{2}}$ represents the magnitude of the inverse square root of variance. And a subnetwork with the parameter of $\bm{\theta}$ is used to model the dependencies among channels by following the designs of the SE block \citep{hu2018squeeze} and GC block \citep{cao2019gcnet}. Here we use a Sigmoid activation to generate a set of channel weights, which is used to control the strength of the inverse square root of variance for each channel. In this way, the output of the IR branch not only has the same magnitude as that of BD but also encodes channel dependencies. We provide detailed descriptions of the subnetwork in Appendix Sec.\ref{sec:subnetwork}.

\subsection{Discussions}\label{sec:discussion}

\textbf{Network Architectures.} 
Different from SE block in Fig.\ref{fig:CE-SE}(a) which only reweights feature channels by a bottleneck network\citep{hu2018squeeze},  CE decorrelates incoming feature channels after the normalization layer by combining two branches, i.e. batch decorrelation (BD) and instance reweighting(IR), as shown in Fig.\ref{fig:CE-SE}(b). 
The CE block can be readily integrated into various advanced architectures, such as ResNet, VGGNet \citep{simonyan2014very}, ShuffleNetv2 \citep{ma2018shufflenet} and MobileNetv2 \citep{sandler2018mobilenetv2}, by inserting it in block of normalization and rectified units.

The flexibility of the CE block makes it easy to construct 
a series of CENets. For example, we consider the residual networks (ResNet). The core unit of the ResNet is the residual block that consists of `$1\times 1$', `$3\times 3$' and `$1\times 1$' convolution layers, sequentially. 
The CE block is applied in the last `$1\times 1$' convolution layer by plugging the CE module before ReLU non-linearity, as shown in Fig.\ref{fig:CE-SE}(c).  Following similar strategies, CE is further integrated into ShuffleNetv2 and MobileNetv2 to construct CE-ShuffleNetv2 and CE-MobileNetv2. whose diagrams are provided in Sec.\ref{sec:netmobile} of Appendix. We also explore the integration strategy used to incorporate CE
blocks into a network architecture in Sec.\ref{sec:appendixD} of Appendix. 

\textbf{Magnitude of Gamma and Feature Channels.}
The CE block can prevent the inhibited channels through the BD branch. 
Remark \ref{remark:1} shows that inhibited channels are usually related to $\gamma_c$ and the output $y_{ncij}$ with small values. 
Here we discover that BD branch can increase the magnitude of gamma and channel features. To see this,
by combining Eqn.(\ref{eq:D_n-relax}) and Eqn.(\ref{eq:CE}), the output of BD can be expressed as  $\pp^{\mathrm{BD}}_{nij}=\Diag(\bm{\Sigma}^{-\frac{1}{2}}\bm{\gamma})\bar{x}_{nij}+\bm{\Sigma}^{-\frac{1}{2}}\bm{\beta}$.
Compared with Eqn.(\ref{eq:bn+relu0}), an equivalent gamma for BD branch can be defined as $\hat{\bm{\gamma}}=\bm{\Sigma}^{-\frac{1}{2}}\bm{\gamma}$. The proposition \ref{prop:equgmma} shows that BD increases the magnitude of $\hat{\bm{\gamma}}$ and feature channels in a feed-forward way. Therefore, it is effective to prevent inhibited channels. The proof of proposition \ref{prop:equgmma} is provided in Sec.\ref{sec:appendixB} of Appendix. 
\vspace{-10pt}
\begin{prop}\label{prop:equgmma}
\vspace{-5pt}
Let $\bm{\Sigma}$ be covariance matrix of feature maps after batch normalization. Then, (1) assume that $\bm{\Sigma}_k=\bm{\Sigma}^{-\frac{1}{2}},\,\forall k=1,2,3,\cdots,T$, we have $|\hat{\gamma_c}|> |\gamma_c|, \, \forall c\in [C]$. (2) Denote $\bm{\rho}=\frac{1}{M}\bar{\bm{x}}\bar{\bm{x}}\tran$ in Eqn.(\ref{eq:batch-co}) and $\tilde{\bm{x}}_{nij}=\Diag(\bm{\gamma})\bar{\bm{x}}_{nij}+\bm{\beta}$. Assume $\bm{\rho}$ is full-rank, then$ \left\|\bm{\Sigma}^{-\frac{1}{2}}\tilde{\bm{x}}_{nij}\right\|_2>\left\|\tilde{\bm{x}}_{nij}\right\|_2$
\vspace{-3pt}
\end{prop}
\vspace{-5pt}

\textbf{Connection with Nash Equilibrium.}
We understand normalization and ReLU block from a perspective in game theory  \citep{leshem2009game}. In this way,
an interesting connection between the proposed CE block and the well-known Nash Equilibrium can be built. To be specific, for every underlying sample, we treat the output $p_{cij}$ in Eqn.(\ref{eq:CE}) as the transmit power allocated to neuron $(i,j)$ for the c-th channel. Here the subscript `$n$' is omitted for clarity. Then each neuron is associated with a maximum information rate which determines the maximum transmit power available to the neuron \citep{cover2012elements}. In strategic games \cite{osborne1994course}, each channel wants to maximize its benefit. In the context of CNN, we suppose that every channel obtains its output by maximizing the sum of the maximum information rate of all neurons.

Furthermore, considering the dependencies among channels, the channels are thought to play a non-cooperative game, named Gaussian interference game, which admits a unique Nash Equilibrium solution \citep{laufer2006game}. When all the outputs are activated (larger than 0), this Nash Equilibrium solution has an explicit expression, the linear proxy of which has the same form with the expression of CE in Eqn.(\ref{eq:CE}). It shows that decorrelating features after the normalization layer can be connected with Nash Equilibrium, implying that the proposed CE block indeed encourages every channel to contribute to the network's computation. Note that the Nash Equilibrium solution can be derived for every single sample, implying that the decorrelation operation should be performed conditioned on each instance sample. This is consistent with our design of the CE block. We present detailed explanations about the connection between CE and Nash Equilibrium in Sec.\ref{sec:appendixC} of the Appendix.

\section{Related Work}
\textbf{Sparsity in ReLU.}
An attractive property of ReLU \citep{sun2015deeply,nair2010rectified} is sparsity, which brings potential advantages such as information disentangling and linear separability. However, \cite{lu2019dying} and \cite{mehta2019implicit} pointed out that some ReLU neurons may become inactive and output 0 values for any input. Previous work tackled this issue by designing new activation functions, such as ELU \citep{clevert2015fast} and  Leaky ReLU \citep{maas2013rectifier}. Recently, \citet{lu2019dying} also tried to solve this problem by modifying the initialization scheme. Different from these work, CE focus on explicitly preventing inhibited channel in a feed-forward way by encouraging channels at the same layer to contribute equally to learned feature representation.

\textbf{Normalization and decorrelation.} There are many practices on normalizer development, such as Batch Normalization (BN) \citep{ioffe2015batch}, Group normalization (GN) \citep{wu2018group} and Switchable Normalization \citep{luo2018differentiable}. A normalization scheme is typically applied after a convolution layer and contains two stages: standardization and rescaling. Another type of normalization methods not only standardizes but also decorrelates features, like DBN \citep{huang2018decorrelated}, IterNorm \citep{huang2019iterative} and switchable whitening \citep{pan2019switchable}. Despite their success in stabilizing the training, little is explored about the relationship between these methods and inhibited channels. Fig.\ref{fig1:sparse-top1first} shows that inhibited channels emerge in VGGNet where `BN+ReLU' or `LN+ReLU' is used. Unlike previous decorrelated normalizations where decorrelation operation is applied after a convolution layer, our CE explicitly decorrelates features after normalization and is designed to prevent inhibited channels emerging in the block of normalization and rectified units.

\section{Experiments}\label{sec:exp}
We extensively evaluate the proposed CE on two basic vision tasks, image classification on ImageNet \citep{russakovsky2015imagenet} and object detection/segmentation on COCO \citep{lin2014microsoft}.

\subsection{Image Classification on ImageNet} \label{imagenetsetting}

We first evaluate CE on the ImageNet benchmark. The models are trained on the $\sim 1.28$M training images and evaluate on the 50,000 validation images. The top-1 and top-5 accuracies are reported.  We are particularly interested in whether the proposed CE has better generalization to testing samples in various modern CNNs such as ResNets \citep{he2016deep}, MobileNetv2 \cite{sandler2018mobilenetv2}, ShuffleNetv2 \citep{ma2018shufflenet} compared with the SE block \citep{hu2018squeeze}. 
The training details are illustrated in Sec.\ref{sec:appendixE} of the Appendix. 

\begin{table*}
\centering
\scriptsize
\begin{tabular}{c|c c c|c c c|c c c}
    \hline
    &\multicolumn{3}{c|}{ResNet18} &\multicolumn{3}{c|}{ResNet50} &\multicolumn{3}{c}{ResNet101}\\
    \hline
    &Baseline &SE &CE &Baseline &SE &CE &Baseline &SE &CE \\
    \hline
    Top-1 &70.4 &71.4 &\textbf{71.9} &76.6 &77.6 &\textbf{78.3} &78.0 &78.5 &\textbf{79.0} \\
    Top-5 &89.4 &90.4 &\textbf{90.8} &93.0 &93.7 &\textbf{94.1} &94.1 &94.1 &\textbf{94.6} \\
    GFLOPs &1.82 &1.82 &1.83 &4.14 &4.15 &4.16 &7.87 &7.88 &7.89 \\
    CPU (s)&3.69&3.69&4.13&8.61&11.08&11.06&15.58&19.34&17.05\\
    GPU (s)&0.003&0.005&0.006&0.005&0.010&0.009&0.011&0.040&0.015\\
    \hline
    \end{tabular}
    \caption{Comparisons with baseline and SENet on ResNet-18, -50, and -101 in terms of accuracy, GFLOPs, CPU and GPU inference time on ImageNet. The top-1,-5 accuracy of our CE-ResNet is higher than SE-ResNet while the computational cost in terms of GFLOPs, GPU and CPU inference time remain nearly the same. }\label{tab:acc-resent}
\hfill
\vspace{-0.4cm}
\end{table*}
\begin{table*}
\centering
\scriptsize
\begin{tabular}{c|c c c|c c c|c c c}
    \hline
    &\multicolumn{3}{c|}{MobileNetv2 \(1\times\)} &\multicolumn{3}{c|}{ShuffleNetv2 \(0.5\times\)}
    &\multicolumn{3}{c}{ShuffleNetv2 \(1\times\)}\\
    \hline
    &top-1 &top-5 &GFLOPs &top-1 &top-5&GFLOPs &top-1 &top-5 &GFLOPs \\
    \hline
    Baseline &72.5 &90.8 &0.33 &59.2 &82.0 &0.05 &69.0 &88.6&0.15 \\
    SE &73.5 &91.7 &0.33 &60.2 &82.4 &0.05 &70.7 &89.6&0.15\\
    CE &\textbf{74.6} &\textbf{91.7} &0.33 &\textbf{60.5} &\textbf{82.7} &0.05&\textbf{71.2} &\textbf{89.8} &0.16 \\
    \hline
    \end{tabular}
    \centering
    \caption{Comparisons with baseline and SE on lightweight networks, MobileNetv2 and ShuffleNetv2, in terms of accuracy and GFLOPs on ImageNet. Our CENet improves the top-1 accuracy by a large margin compared with SENet with nearly the same GFLOPs.}\label{tab:acc-mobile}
\hfill
\vspace{-0.1cm}
\end{table*}

\textbf{Performance comparison on ResNets.} We evaluate CE on representative residual network structures including ResNet18, ResNet50 and ResNet101. The CE-ResNet is compared with baseline (plain ResNet) and SE-ResNet. For fair comparisons, we use publicly available code and re-implement baseline models and SE modules with their respective best settings in a unified Pytorch framework. To save computation, the CE blocks are selectively inserted into the last normalization layer of each residual block. Specifically, for ResNet18, we plug the CE block into each residual block. For ResNet50, CE is inserted into all residual blocks except for those layers with 2048 channels. For ResNet101, the CE blocks are employed in the first seven residual blocks.
%
%
%

\textbf{Improved generalization on ResNets.} As shown in Table \ref{tab:acc-resent}, our proposed CE outperforms the BN baseline and SE block by a large margin with little increase of GFLOPs. Concretely, CE-ResNet18, CE-ResNet50 and CE-ResNet101 obtain top-1 accuracy increase of $1.5\%$, $1.7\%$ and $1.0\%$ compared with the corresponding plain ResNet architectures, confirming the improved generalization on testing samples. Note that the shallower
network, i.e. CE-ResNet50, even outperforms the deeper network, i.e. plain ResNet101 (78.0),  suggesting that the learned features under CE blocks are more representative. We plot training and validation error during the training process for ResNet50, SE-ResNet50 and CE-ResNet50 in Fig.\ref{fig:CEeuqal}(a). Compared to ResNet50 and SE-ResNet50, CE-ResNet50 obtains lower training error and validation error than that of SE-ResNet50, implying that CE improves the generalization ability of the network. 

\textbf{Comparable computational cost.} We also analyze the complexity of BN, SE, and CE in terms of GFLOPs, GPU and CPU running time. The definition of GFLOPs
follows \citep{sandler2018mobilenetv2}, $i.e$., the number of multiply-adds. We evaluate the inference time\footnote{The CPU type is Intel Xeon CPU E5-2682 v4, and the GPU is NVIDIA GTX1080TI. The implementation is based on Pytorch} with a mini-batch of 32. In terms of GFLOPs, the CE-ResNet18, CE-ResNet50, CE-ResNet101 has only $0.55\%$, $0.48\%$ and $0.25\%$ relative increase in GFLOPs compared with plain ResNet. Additionally, the CPU and GPU inference time of CENet is nearly the same with SENet.

\textbf{Improved generalization  on light-weight networks}.
We further investigate the efficacy of our proposed CE in two representative light-weight networks, MobileNetv2 and ShuffleNetv2. The results of the comparison are given in Table \ref{tab:acc-mobile}. It is seen that CE blocks bring conspicuous improvements in top-1 and top-5 accuracies on test examples at a minimal increase in computational burden. For MobileNetv2 \(1\times\), CE  even improves top-1 accuracy of baseline by $2.1\%$, showing that CE enables the network to generalize well in testing samples.

\subsection{Analysis of CE}\label{sec:ce-analysis}
In this section, we investigate the robustness of CE against label corruptions \citep{zhang2016understanding}. We demonstrate that CE encourages channels to contribute equally to the learned feature representation and reduces correlations among feature channels. More experimental results are presented in Appendix Sec.\ref{sec:appendixD}.

\begin{figure}
\centering
\begin{subfigure}{.4\columnwidth}
\centering
\includegraphics[width=1.0\textwidth]{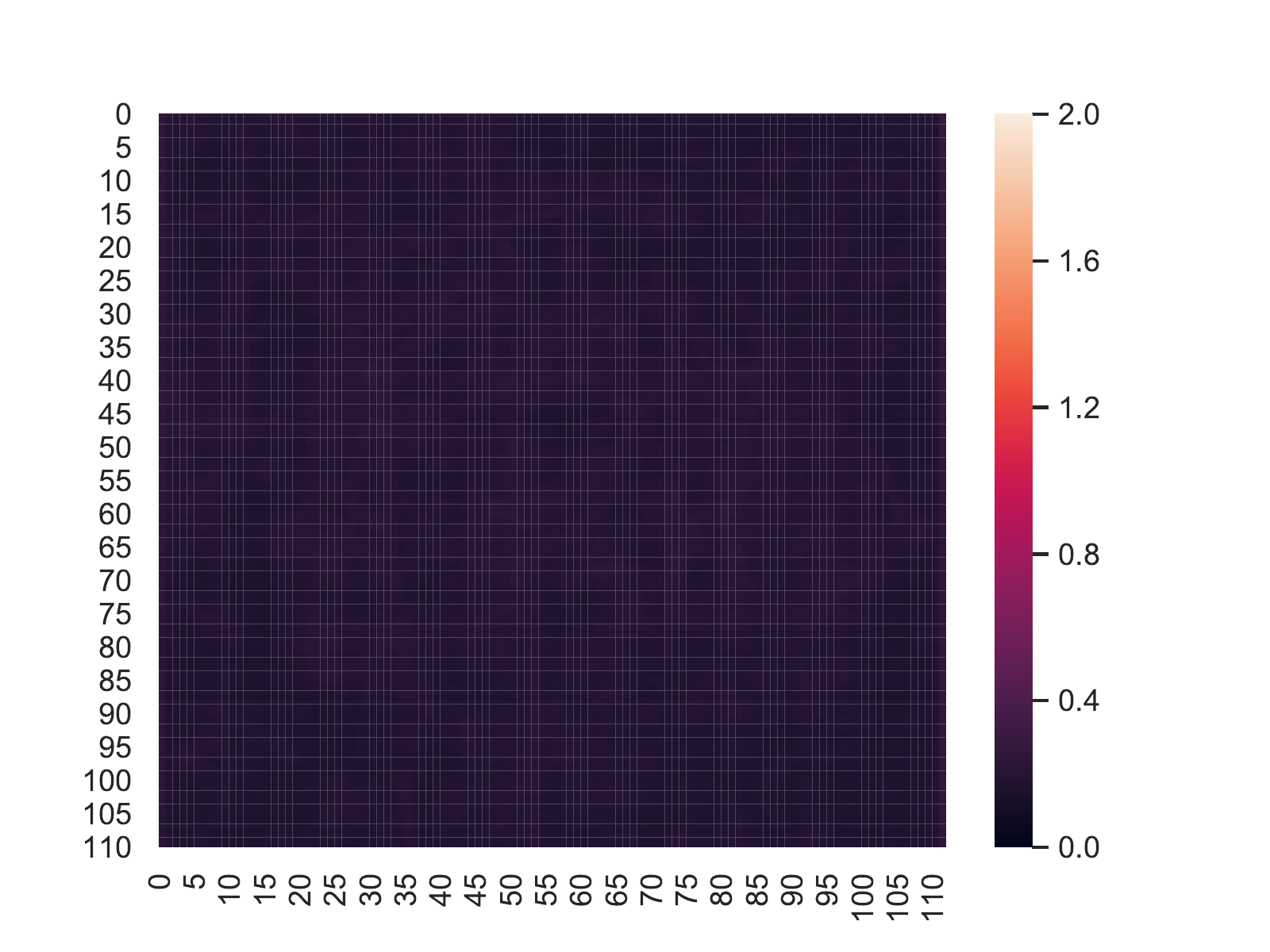}
\caption*{}
\end{subfigure}
\begin{subfigure}{.4\columnwidth}
\centering
\includegraphics[width=1.0\textwidth]{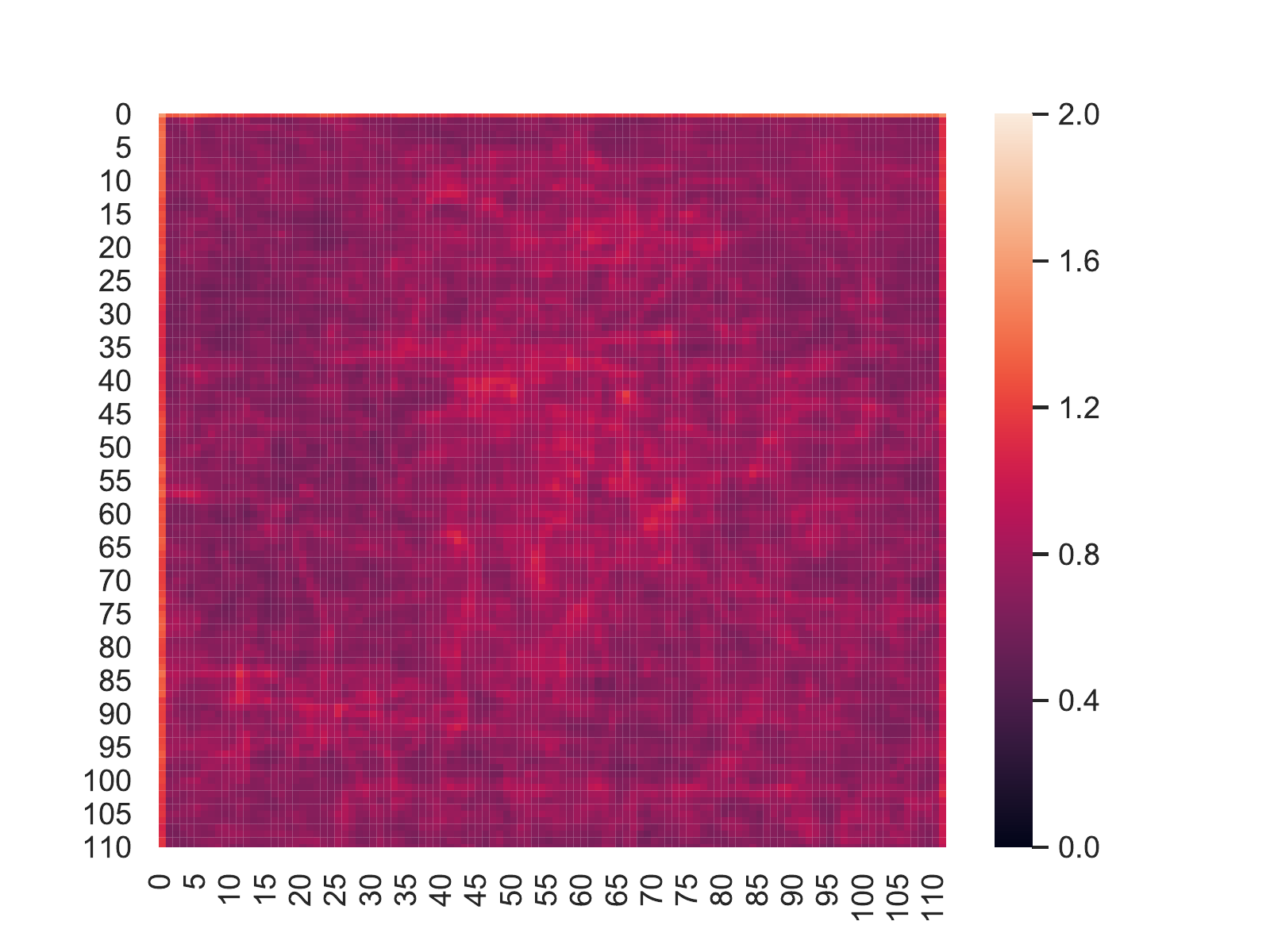}
\caption*{}
\end{subfigure}
\vspace{-0.5cm}
\caption{\textbf{Left} {\&} \textbf{Right} show the magnitude of feature channels after BN and CE layer, respectively. The $\ell_2$ norm of feature channels at each location $(i,j)$ after the first BN or CE layer of the trained VGGNet is visualized. CE increase the magnitude of channel features.}
\label{fig:CEplot}
\vspace{-0.5cm}
\end{figure}

\begin{figure*}
\centering
\begin{subfigure}{.4\columnwidth}
\centering
\includegraphics[width=1.0\textwidth]{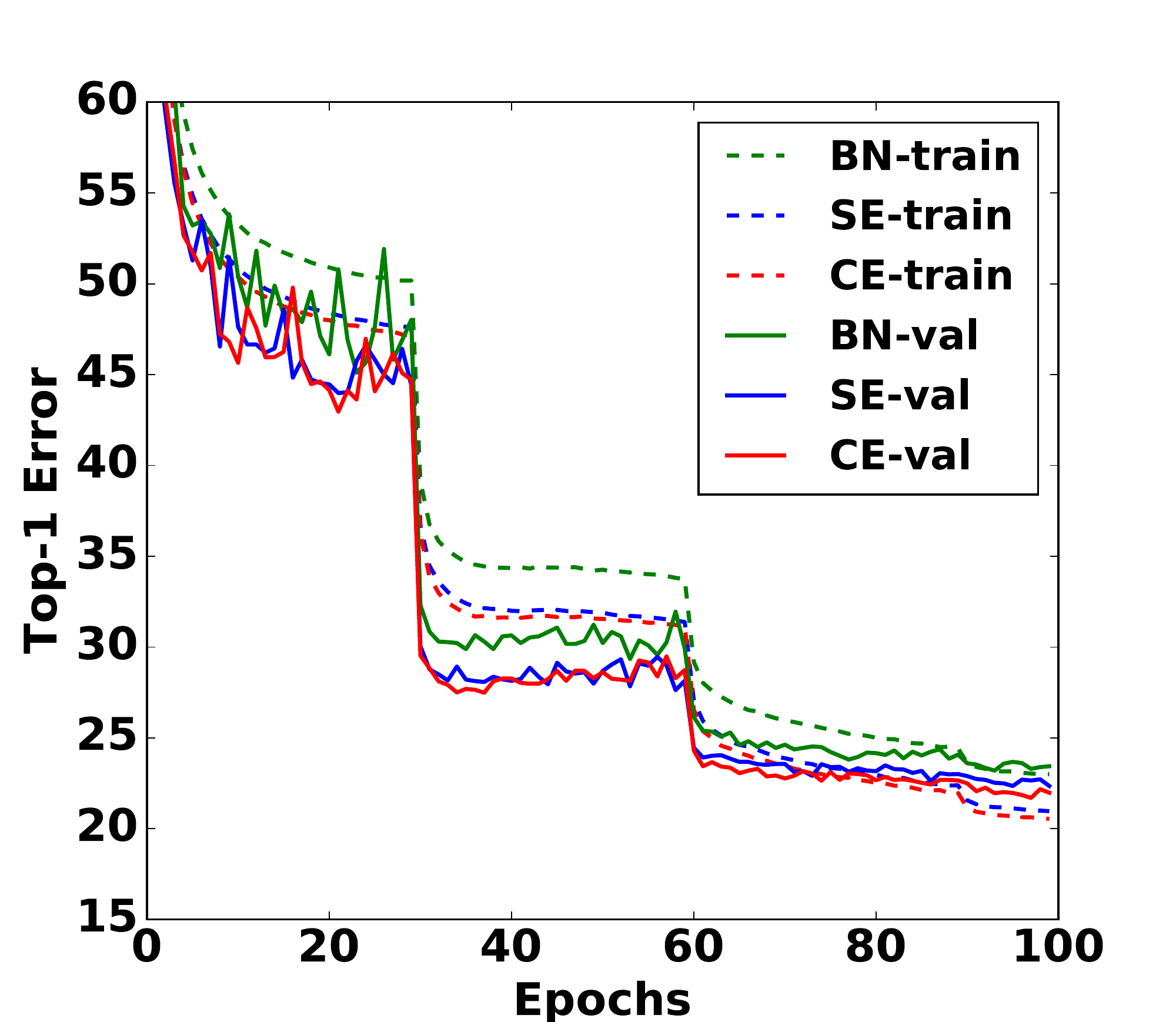}
\caption{}
\end{subfigure}
\begin{subfigure}{.4\columnwidth}
\centering
\includegraphics[width=0.96\textwidth]{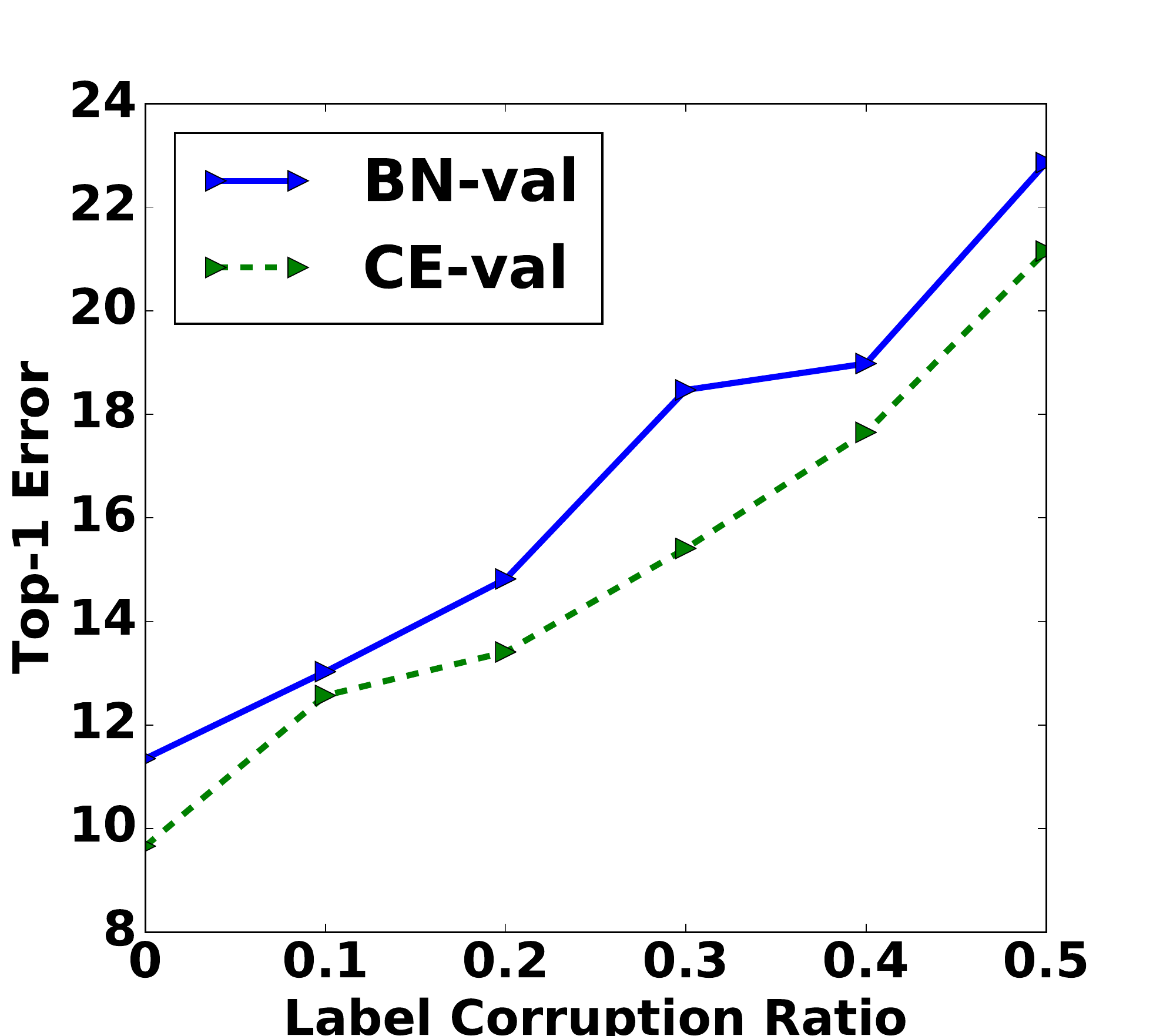}
\caption{}
\end{subfigure}
\begin{subfigure}{.4\columnwidth}
\centering
\includegraphics[width=1.0\columnwidth]{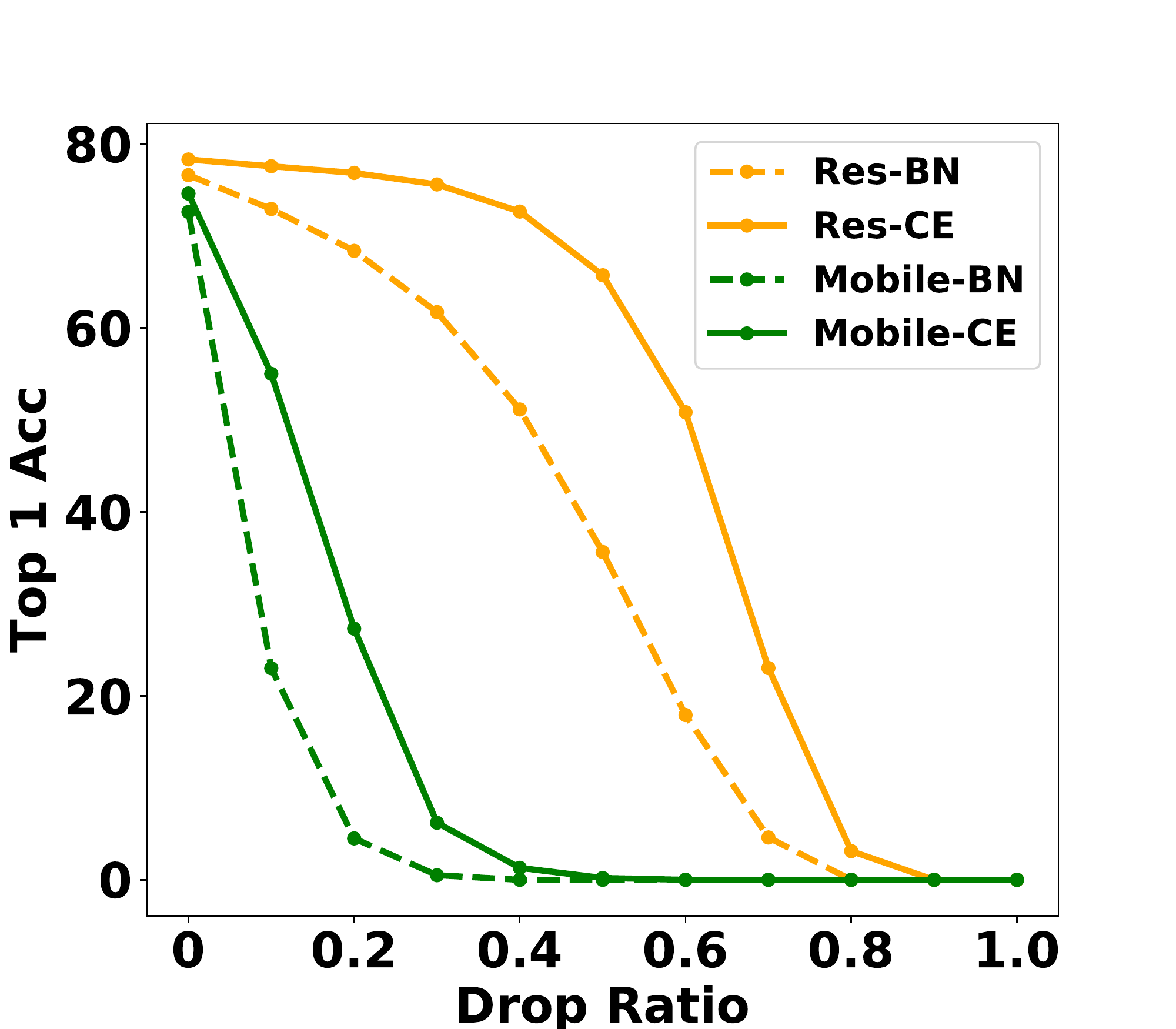}
\caption{}
\end{subfigure}
\begin{subfigure}{.4\columnwidth}
\centering
\includegraphics[width=0.95\textwidth]{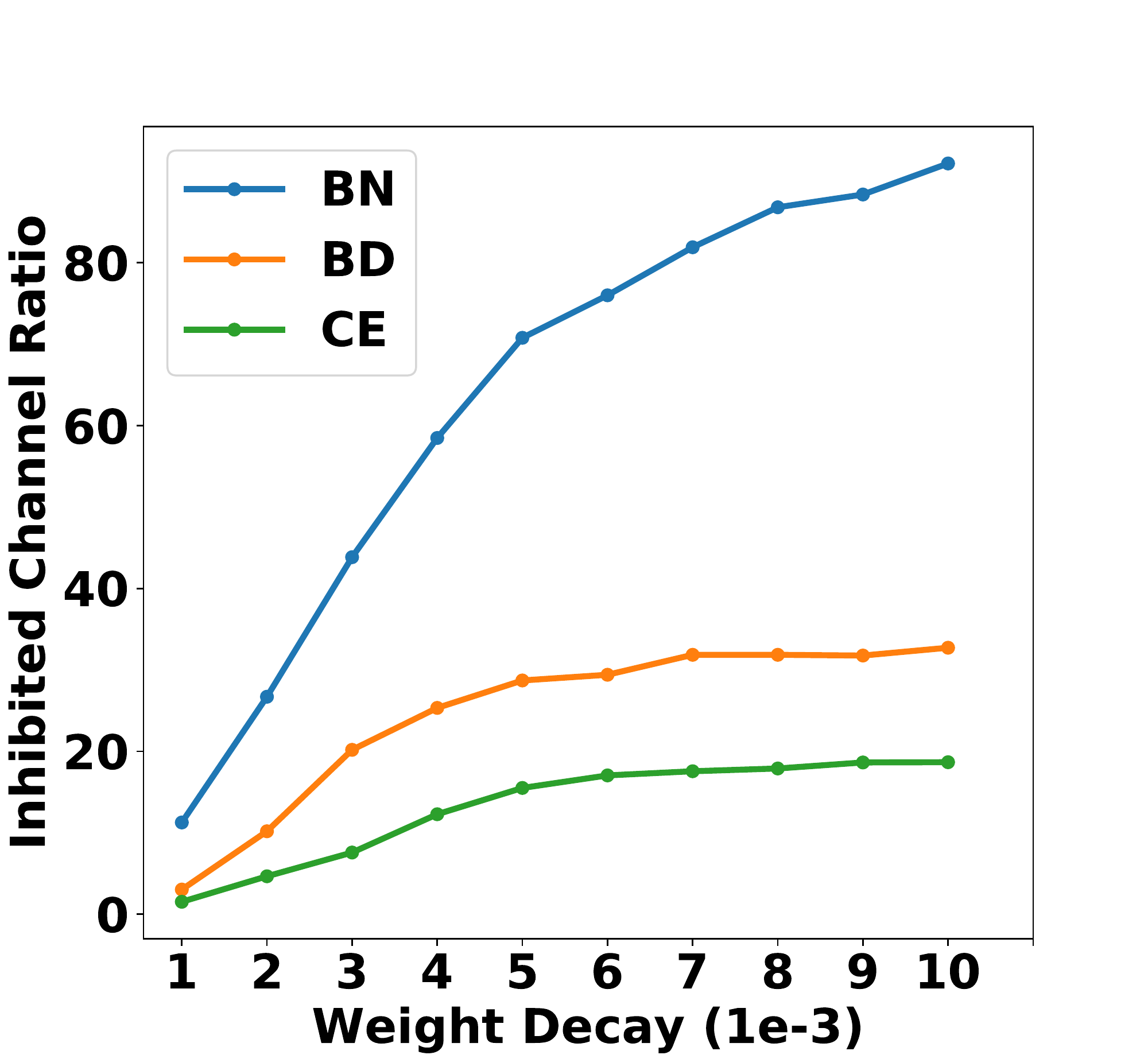}
\caption{}
\end{subfigure}
\begin{subfigure}{.4\columnwidth}
\centering
\includegraphics[width=1.0\textwidth]{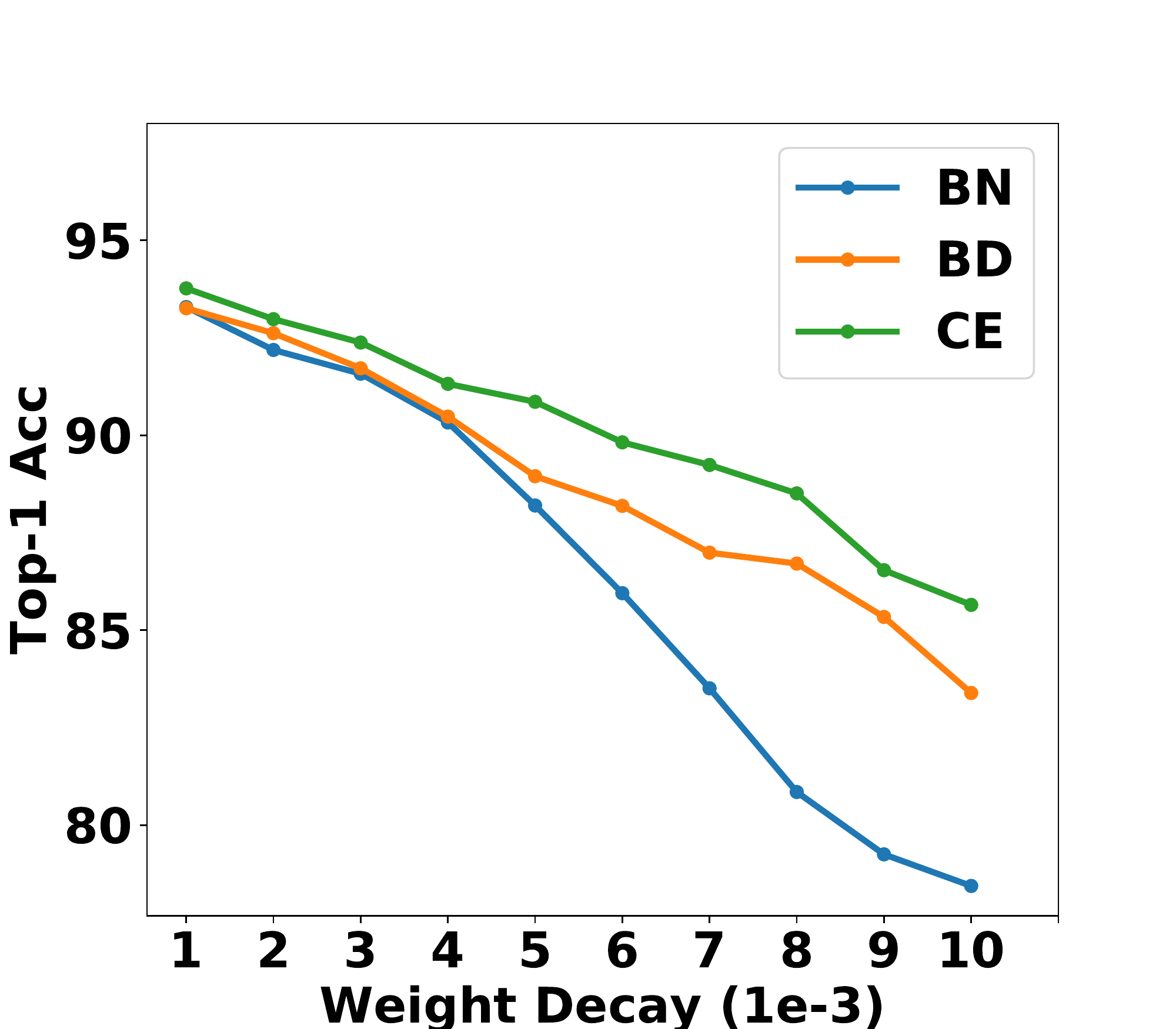}
\caption{}
\end{subfigure}
\vspace{-0.3cm}
\caption{(a) shows the training and validation error curves on ImageNet with ResNet50 as backbone for BN, SE and CE. CE improves both training error and validation error. (b) shows
Robustness test of CE at 5-level corruption labels on CIFAR10 dataset where `0' corruption indicates no corrupted labels. 
CE gives improved test error over baselines in all label corruption ratios. (c) shows cumulative ablation curves for MobileNetv2 and ResNet50 on ImageNet dataset respectively. We randomly ablate channels with an increasing fraction in the first normalization layers. CE also helps to equalize the importance of channels on ImageNet. (d) \& (e) are inhibited channel ratio and top-1 accuracy curves when training VGGNet on CIFAR-10 under different weight decays. Compared to networks trained with BN, networks trained with the proposed BD and CE can effectively prevent inhibited channels and retain a higher performance as strength of weight decay increases.}
\label{fig:CEeuqal}
\vspace{-0.3cm}
\end{figure*}

\textbf{CE improves generalization ability in corrupted label setting.} We have shown in Sec.\ref{imagenetsetting} that CE has a better generalization to testing samples that are drawn from the same distributions of training ones. Here we show the robustness of CE when the labels of training samples are randomly corrupted with different corruption ratios \cite{zhang2016understanding}. We train VGGNet with BN and CE on CIFAR10 \citep{krizhevsky2009learning} under the same training settings in Fig.\ref{fig1:sparse-top1first}. Especially,  VGGNet with CE is trained to obtain the same training error of VGGNet with BN. The top-1 test errors of VGGNet with BN and CE are plotted in  Fig.\ref{fig:CEeuqal}(e). It shows that CE consistently improves the generalization ability under a wide range of corruption label ratios.

%

\textbf{CE encourages channels to contribute more equally to the learned feature representation.}  
We demonstrate this in two ways. First, by applying a decorrelation operator, neurons across $C$ channels after CE block have a larger magnitude at every location $(i,j)$. We use $\ell_2$ norm to measure the magnitude of feature channels. The average of the magnitude for each location $(i,j)$ after CE blocks are calculated over a random minibatch of samples. Results are obtained by training BN-VGGNet and CE-VGGNet. Fig.\ref{fig:CEplot} shows that neurons across channels in CE-VGGNet have a larger magnitude than those in BN-VGGNet, meaning that CE makes more channels useful in the feature representation.

Second, the importance of feature channels to the network's prediction is more equal. We investigate this by using a cumulative ablation method \citep{morcos2018importance}. Typically, the importance of a single channel to the network's computation can be measured by the relative performance drop once that channel is removed (clamping activity a feature map to zero). If the importance of channels to the network's prediction is more equal, the network will rely less on some specific channels and thus the performance will drop more gently. With this method, we see how ResNet50 and MobileNetv2 $1\times$ with CE blocks respond to the cumulative random ablation of channels on ImageNet. We plot the ablation ratio versus the top-1 accuracy in Fig.\ref{fig:CEeuqal}(c). It can be observed that the CE block is able to resist the cumulative random ablation of channels on both ResNet50 and MobileNetv2 compared with the original networks, showing that CE can effectively make channels contribute more equally to the network's prediction.

\textbf{CE mitigates the inhibited channels, which is robust to different strength of weight decay.} 
\citep{mehta2019implicit} revealed that the number of inhibited channels increases as the strength of weight decay grows. As shown in Fig.\ref{fig:CEeuqal}(d), the number of inhibited channel in CE-VGGNet trained on CIFAR10 is conspicuously reduced under all weight decays compared with BN-VGGNet. We also note that the BD branch in the CE block is also able to prevent inhibited channels, which is consistent with proposition \ref{prop:equgmma}. CE achieves the lower inhibited channel ratio than BD, implying that IR also helps to prevent inhibited channel. Fig.\ref{fig:CEeuqal}(e) further shows that the top-1 accuracy of VGGNet with BN drops significantly as the weight decay increases, but CE can alleviate accuracy drop, implying that excessive inhibited channels impede network' generalization to testing samples.
\begin{figure}
\centering
\begin{subfigure}{.32\columnwidth}
\centering
\includegraphics[width=1.0\textwidth]{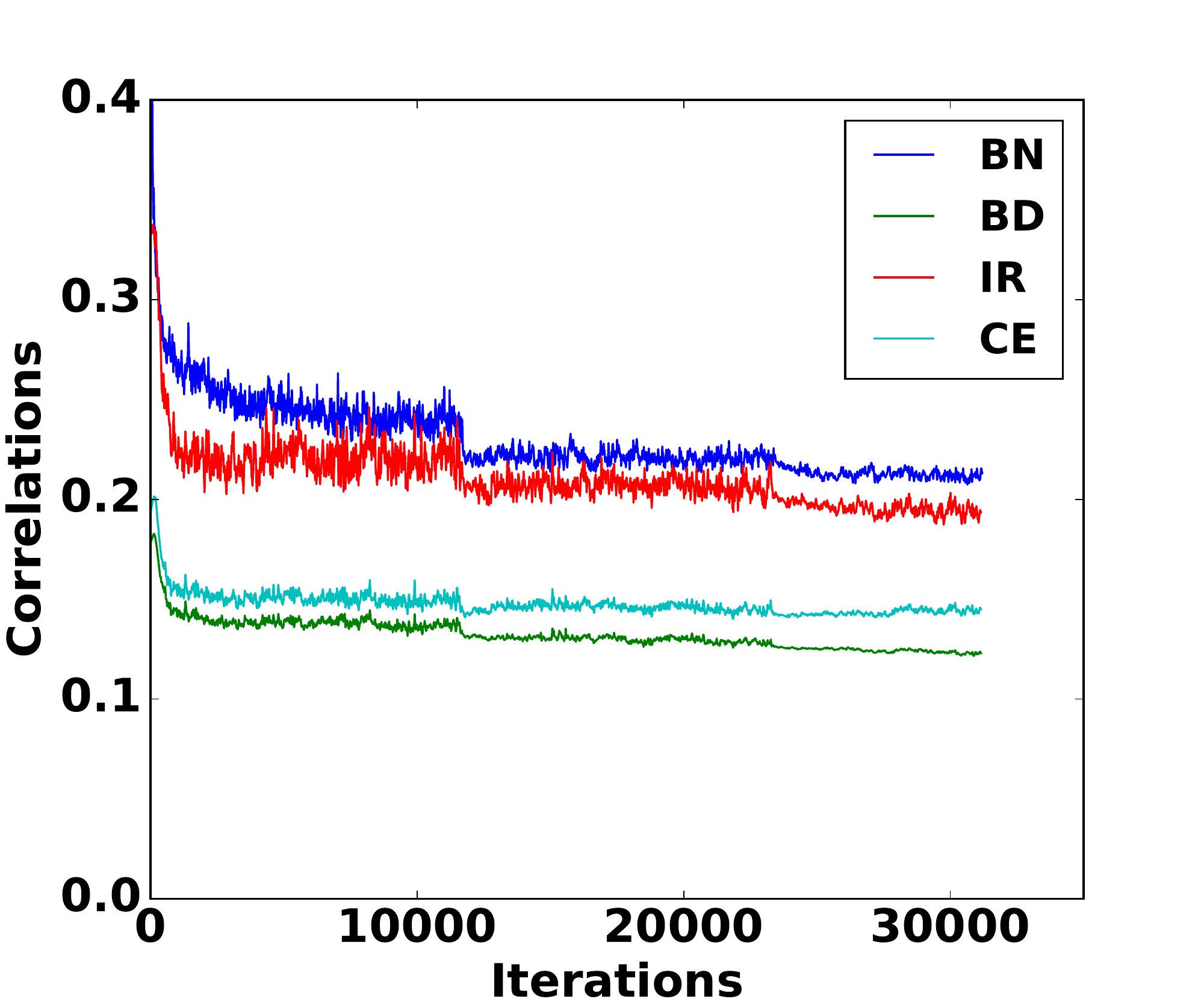}
\caption*{}
\end{subfigure}
\begin{subfigure}{.32\columnwidth}
\centering
\includegraphics[width=1.0\textwidth]{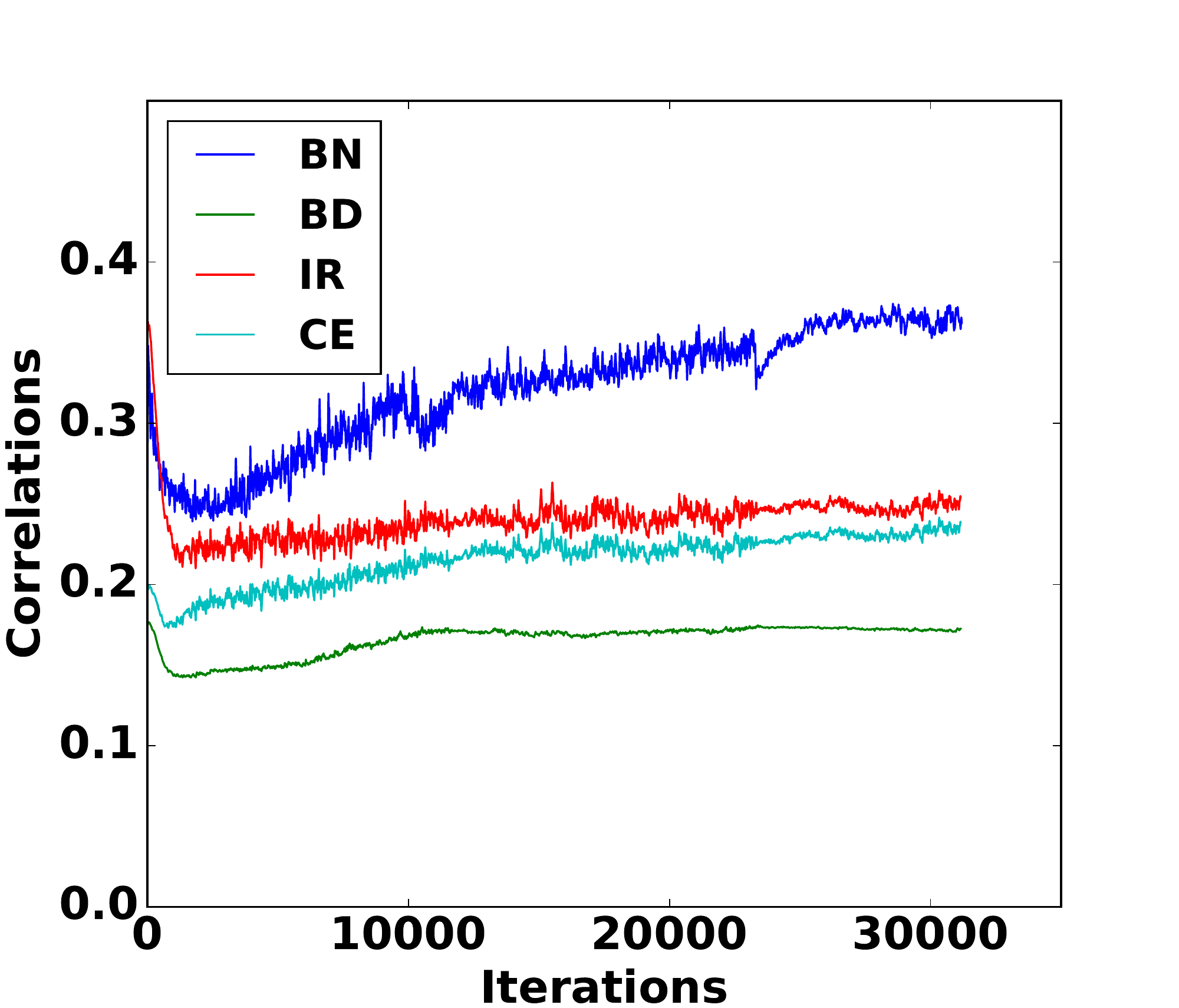}
\caption*{}
\end{subfigure}
\begin{subfigure}{.32\columnwidth}
\centering
\includegraphics[width=0.95\textwidth]{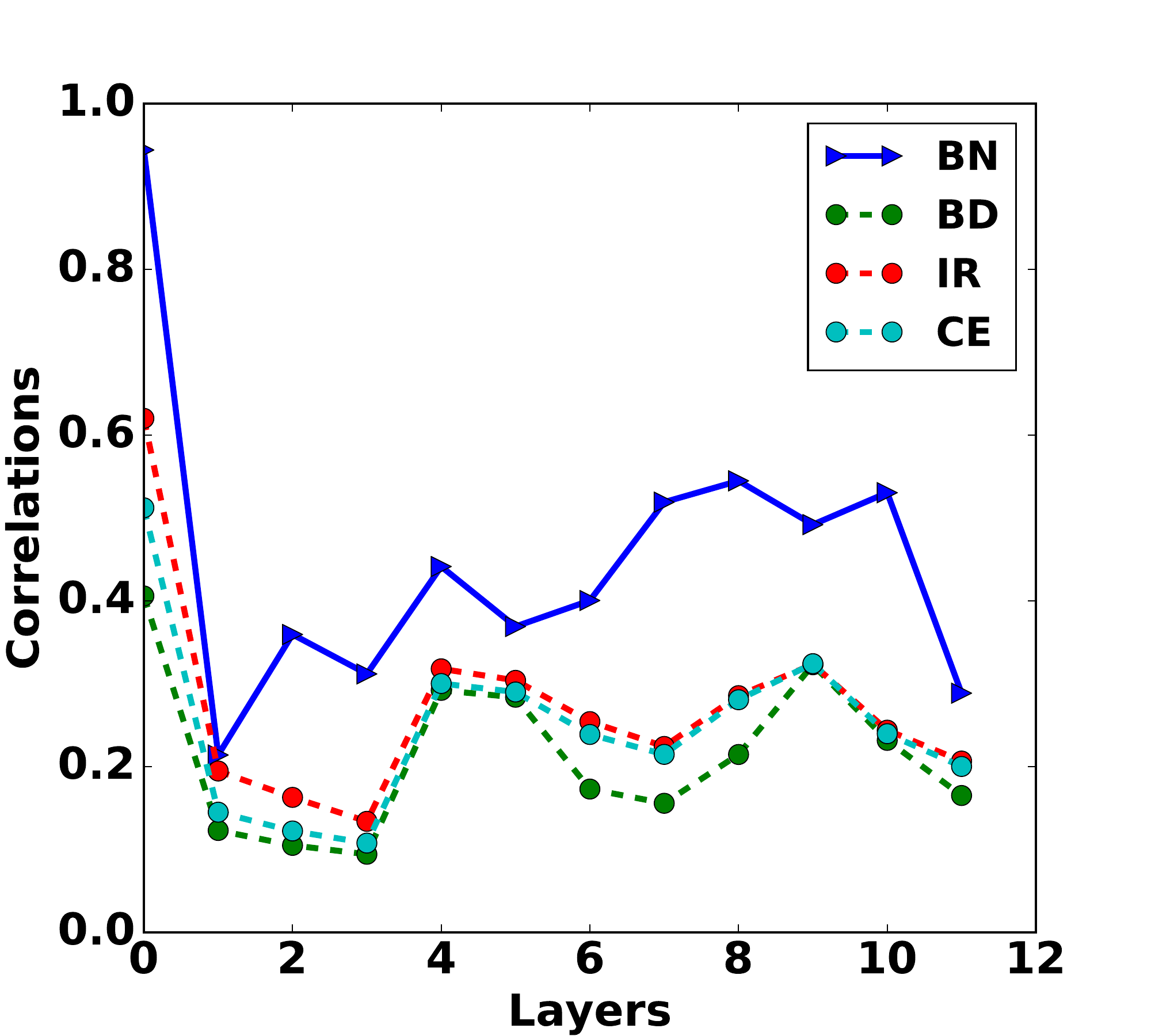}
\caption*{}
\end{subfigure}
\vspace{-7pt}
\caption{\textbf{Left} \& \textbf{Middle} show the correlations for the output response maps in shallow (Left) and deeper (Middle) CE layers during the whole training period. \textbf{Right} shows the curves of correlations at different layers. Results are obtained by applying VGGNet as backbone. All of CE, IR and BD can achieve lower correlations among feature channels than BN baseline.}
\label{fig:CE-corr}
\vspace{-3pt}
\end{figure}

\vspace{-6pt}
\textbf{CE reduces correlations among feature channels.} By design, CE  decorrelates feature channels by the BD branch, which is then used to generate a decorrelation operator conditioned on each sample by combining with the IR branch. We investigate the effect of reducing correlations among feature channels of BN (Baseline), IR, BD, and CE by applying VGGNet. As shown in Fig.\ref{fig:CE-corr}, the correlations among feature channels at different depths of the network are remarkably decreased when CE, IR, and BD are used, implying that CE can reduce the redundancy in feature channels. We also observe that CE can learn adaptive correlations at different depths of the network by combining BD and IR. Note that in deeper layers of the network, the decorrelation of CE behaves more similar to decorrelation of IR compared with BD, showing that decorrelating feature channels for each instance is useful in higher layers.

\subsection{Object Detection and Instance Segmentation on COCO}
We assess the generalization of our CE block on detection/segmentation track using the COCO2017 dataset ~\citep{lin2014microsoft}. We train our model on the union of 80k training images and 35k validation images and report the performance on the mini-val 5k images. Mask-RCNN is used as the base detection/segmentation framework. The standard COCO metrics of Average Precision (AP) for bounding box detection (APbb) and instance segmentation (APm) is used to evaluate our methods. 
In addition, we adopt two common training settings for our models, (1) freezing the vanilla BN and CE layer and (2) updating parameters with the synchronized version. 
For vanilla BN and CE layers, all the gamma, beta parameters, and the tracked running statistics are frozen.
In contrast, for the synchronized version, the running mean and variance for batch normalization and the covariance for CE layers are computed across multiple GPUs.
The gamma and beta parameters are updated during training while $\tilde{F}$ and $\lambda$ are frozen to prevent overfitting.
We use MMDetection training framework with ResNet50/ResNet101 as basic backbones and all the hyper-parameters are the same as \cite{chen2019mmdetection}. Table \ref{coco} shows the detection and segmentation results. The results show that compared with vanilla BN, our CE block can consistently improve the performance. For example, our fine-tuned CE-ResNet50 is 2.2 AP higher in detection and 2.7 AP higher in segmentation. For the sync BD version, CE-ResNet50 gets \textbf{42.0} AP in detection and \textbf{37.5} AP in segmentation, which is the best performance for ResNet50 to the best of our knowledge. To sum up, these experiments demonstrate the generalization ability of CE blocks in other tasks.

\begin{table}[]
    \centering
    \scriptsize
    \begin{tabular}{c|c c c|c c c}
    \hline
         Backbone&\(AP^b\)&\(AP^b_{.5}\)&\(AP^b_{.75}\)&\(AP^m\)&\(AP^m_{.5}\)&\(AP^m_{.75}\)  \\
    \hline
    \hline
         ResNet50&38.6&59.5&41.9&34.2&56.2&36.1\\
         +CE &40.8&\textbf{62.7}&44.3&36.9&59.2&39.4\\
         +SyncCE &\textbf{42.0}&62.6&\textbf{46.1}&\textbf{37.5}&\textbf{59.5}&\textbf{40.3}\\
         \hline
    \hline
    ResNet101&40.3&61.5&44.1&36.5&58.1&39.1\\
    +CE&\textbf{41.6}&\textbf{62.8}&\textbf{45.8}&\textbf{37.4}&\textbf{59.4}&\textbf{40.0}\\
    \hline
    \end{tabular}
    \vspace{-0.1cm}
    \caption{Detection and segmentation results in COCO using Mask-RCNN We use the pretrained CE-ResNet50 model (78.3) and CE-ResNet101 (79.0) in ImageNet to train our model. CENet can consistently improve both box AP and segmentation AP by a large margin.}
    \label{coco}
    \vspace{-0.3cm}
\end{table}
\vspace{-0.3cm}
\section{Conclusion}
In this paper, we presented an effective and efficient network block, termed as Channel Equilibrium (CE). We show that CE encourages channels at the same layer to contribute equally to learned feature representation, enhancing the generalization ability of the network. Specifically, CE can be stacked between the normalization layer and the Rectified units, making it
flexible to be integrated into various CNN architectures. The superiority of CE blocks has been demonstrated on the task of image classification and instance segmentation. We hope that the analyses of CE could bring a new perspective for future work in architecture design.

\bibliography{iclr2020_conference}
\bibliographystyle{icml2020}

\clearpage

\onecolumn

\appendix{\Large{Appendix}}

\section{Proof of Remark \ref{remark:1} }\label{sec:appendixremark1}
 Let $z\sim \mathcal{N}(0,1)$  and $y=\max\{0,\gamma_cz+\beta_c\}$. For the suffiency, when $\gamma_c>0$ we have
\begin{equation}\label{eq:limgamma1}
\begin{split}
\mathbb{E}_z[y]&=\int_{-\infty}^{-\frac{\beta_c}{\gamma_c}}0\cdot \frac{1}{\sqrt{2\pi}}\mathrm{exp}^{-\frac{z^2}{2}}dz+\int_{-\frac{\beta_c}{\gamma_c}}^{+\infty}(\gamma_cz+\beta_c)\cdot \frac{1}{\sqrt{2\pi}}\mathrm{exp}^{-\frac{z^2}{2}}dz,\\
&=\frac{\gamma_c\mathrm{exp}^{-\frac{\beta_c^2}{2\gamma_c^2}}}{\sqrt{2\pi}}+\frac{\beta_c}{2}(1+\mathrm{Erf}[\frac{\beta_c}{\sqrt{2}\gamma_c}]),
\end{split}
\end{equation}
where $\mathrm{Erf}[x]=\frac{2}{\sqrt{\pi}}\int_0^{x}\mathrm{exp}^{-t^2}dt$ is the error function. From Eqn.(\ref{eq:limgamma1}), we have
\begin{equation}\label{eq:limgamma2}
\begin{split}
\lim_{\gamma_c\rightarrow 0^+}\mathbb{E}_z[y]&=\lim_{\gamma_c\rightarrow 0^+}\frac{\gamma_c\mathrm{exp}^{-\frac{\beta_c^2}{2\gamma_c^2}}}{\sqrt{2\pi}}+\lim_{\gamma_c\rightarrow 0^+}\frac{\beta_c}{2}(1+\mathrm{Erf}[\frac{\beta_c}{\sqrt{2}\gamma_c}])=0
\end{split}
\end{equation}
In the same way, we can calculate
\begin{equation}\label{eq:limgamma3}
\begin{split}
\mathbb{E}_z[y^2]&=\int_{-\infty}^{-\frac{\beta_c}{\gamma_c}}0\cdot \frac{1}{\sqrt{2\pi}}\mathrm{exp}^{-\frac{z^2}{2}}dz+\int_{-\frac{\beta_c}{\gamma_c}}^{+\infty}(\gamma_cz+\beta_c)^2\cdot \frac{1}{\sqrt{2\pi}}\mathrm{exp}^{-\frac{z^2}{2}}dz,\\
&=\frac{\gamma_c\beta_c\mathrm{exp}^{-\frac{\beta_c^2}{2\gamma_c^2}}}{\sqrt{2\pi}}+\frac{\gamma_c^2+\beta_c^2}{2}(1+\mathrm{Erf}[\frac{\beta_c}{\sqrt{2}\gamma_c}]),
\end{split}
\end{equation}
From Eqn.(\ref{eq:limgamma3}), we have
\begin{equation}\label{eq:limgamma4}
\begin{split}
\lim_{\gamma_c\rightarrow 0^-}\mathbb{E}_z[y^2]&=\lim_{\gamma_c\rightarrow 0^+}\frac{\gamma_c\beta_c\mathrm{exp}^{-\frac{\beta_c^2}{2\gamma_c^2}}}{\sqrt{2\pi}}+\lim_{\gamma_c\rightarrow 0^+}\frac{\gamma_c^2+\beta_c^2}{2}(1+\mathrm{Erf}[\frac{\beta_c}{\sqrt{2}\gamma_c}])=0
\end{split}
\end{equation}
When $\gamma_c<0$, we have
\begin{equation}\label{eq:limgamma5}
\mathbb{E}_z[y]=-\frac{\gamma_c\mathrm{exp}^{-\frac{\beta_c^2}{2\gamma_c^2}}}{\sqrt{2\pi}}+\frac{\beta_c}{2}(1-\mathrm{Erf}[\frac{\beta_c}{\sqrt{2}\gamma_c}]),
\end{equation}
and
\begin{equation}\label{eq:limgamma6}
\mathbb{E}_z[y^2]=-\frac{\gamma_c\beta_c\mathrm{exp}^{-\frac{\beta_c^2}{2\gamma_c^2}}}{\sqrt{2\pi}}+\frac{\gamma_c^2+\beta_c^2}{2}(1-\mathrm{Erf}[\frac{\beta_c}{\sqrt{2}\gamma_c}]),
\end{equation}
If $\gamma_c$ sufficiently approaches 0, we arrive at
\begin{equation}\label{eq:limgamma7}
\lim_{\gamma_c\rightarrow 0^+}\mathbb{E}_z[y]=\lim_{\gamma_c\rightarrow 0^-}-\frac{\gamma_c\mathrm{exp}^{-\frac{\beta_c^2}{2\gamma_c^2}}}{\sqrt{2\pi}}+\lim_{\gamma_c\rightarrow 0^-}\frac{\beta_c}{2}(1-\mathrm{Erf}[\frac{\beta_c}{\sqrt{2}\gamma_c}])=0
\end{equation}
and
\begin{equation}\label{eq:limgamma8}
\begin{split}
\lim_{\gamma_c\rightarrow 0^-}\mathbb{E}_z[y^2]&=\lim_{\gamma_c\rightarrow 0^-}\frac{-\gamma_c\beta_c\mathrm{exp}^{-\frac{\beta_c^2}{2\gamma_c^2}}}{\sqrt{2\pi}}+\lim_{\gamma_c\rightarrow 0^-}\frac{\gamma_c^2+\beta_c^2}{2}(1+\mathrm{Erf}[\frac{\beta_c}{\sqrt{2}\gamma_c}])=0
\end{split}
\end{equation}

For necessity, we show that if $\mathbb{E}_z[y]=0$ and $\mathbb{E}_z[y^2]=0$, then $\gamma_c \rightarrow 0$ and $\beta_c\leq 0$. First, if $\gamma_c>0$, combining Eqn Eqn.(\ref{eq:limgamma1}) and Eqn.(\ref{eq:limgamma3}) gives us $\gamma_c \rightarrow 0$ and $\beta_c\leq 0$.
If $\gamma_c<0$, combining Eqn Eqn.(\ref{eq:limgamma5}) and Eqn.(\ref{eq:limgamma6}), we can also obtain $\gamma_c \rightarrow 0$ and $\beta_c\leq 0$. This completes the proof.

 Note that Eqn.(\ref{eq:limgamma2}) and Eqn.(\ref{eq:limgamma4}) are obtained by assuming that $\gamma_c\rightarrow 0$ and $\beta_c\leq 0$. The first condition was verified by \citep{mehta2019implicit} that showed that inhibited channels and gamma with small values would emerge at the same time. Here,  We evaluate the second assumption in various ResNets trained on the ImageNet dataset. The percentage of  $\beta_c \leq 0$ in BN after training are reported in Table \ref{tab:gammabeta}. We see that a large amount of $\beta_c$ is non-positive.
 
\begin{table}[H]
    \centering
    \small
\begin{tabular}{c c c c}
    \hline
         CNNs &ResNet18 &ResNet50 &ResNet101\\
    \hline
         ($\beta_c\leq 0$) &76.0 &76.7 &81.8\\
    \hline
    \end{tabular}
    \vspace{-0.1cm}
    \caption{Ratios of ($\beta_c \leq 0$) after traing on various CNNs. }
    \label{tab:gammabeta}
    \vspace{-0.1cm}
\end{table}

\section{Computation details in 'BN-CE-ReLU' block}\label{sec:appendixA}
As discussed before, CE processes incoming features after normalization layer by combining two branches, $i.e.$, batch decorrelation and instance reweighting. 
The former computes a covariance matrix and the latter calculates instance variance. We now take 'BN-CE-ReLU' block as an example to show the computation details of statistics in CE. Given a tensor $\bm{x}\in\R^{N\times C\times H\times W}$, the mean and variance in $\IN$ \citep{ulyanov2016instance} are calculated as:
\begin{equation}\label{eq:IN}
    (\bm{\mu}_{\IN})_{nc}=\frac{1}{HW}\sum^{H,W}_{i,j}x_{ncij},\quad (\bm{\sigma}^2_{\IN})_{nc}=\frac{1}{HW}\sum^{H,W}_{i,j}(x_{ncij}-(\bm{\mu}_{\IN})_{nc})^2
\end{equation}
Hence, we have $\bm{\mu}_{\IN}, \bm{\sigma}^2_{\IN}\in\R^{N\times C}$. Then, the statistics in BN can be reformulated as follows:
\begin{equation}\label{eq:BN}
\begin{split}
    (\bm{\mu}_{\BN})_{c}&=\frac{1}{NHW}\sum^{N,H,W}_{n,i,j}x_{ncij}=\frac{1}{N}\sum_{i}^N\frac{1}{HW}\sum^{H,W}_{i,j}x_{ncij}\\
    (\bm{\sigma}^2_{\BN})_{c}&=\frac{1}{NHW}\sum^{N,H,W}_{n,i,j}(x_{ncij}-(\bm{\mu}_{\BN})_{c})^2\\
    &=\frac{1}{N}\sum_n^{N}\frac{1}{HW}\sum_{i,j}^{H,W}(x_{ncij}-(\bm{\mu}_{\IN})_{nc}+(\bm{\mu}_{\IN})_{nc}-(\bm{\mu}_{\BN})_{c})^2\\
    &=\frac{1}{N}\sum_n^{N}(\frac{1}{HW}\sum_{i,j}^{H,W}(x_{ncij}-(\bm{\mu}_{\IN})_{nc})^2+((\bm{\mu}_{\IN})_{nc}-(\bm{\mu}_{\BN})_{c})^2)\\
    &=\frac{1}{N}\sum_n^{N}(\bm{\sigma}^2_{\IN})_{nc}+\frac{1}{N}\sum_n^{N}((\bm{\mu}_{\IN})_{nc}-(\bm{\mu}_{\BN})_{c})^2
\end{split}
\end{equation}
Then, we have $\mu_{\BN}=\Expe[\mu_{\IN}]$ and $\sigma^2_{\BN}=\Expe[\sigma^2_{\IN}]+\Dvar[\mu_{\IN}]$, where $\Expe[\cdot]$ and $\Dvar[\cdot]$ denote expectation and variance operators over N samples. Further, the input of IR is instance variance of features estimated by $\tilde{\bm{x}}$, which can be calculated as follows:
\begin{equation}\label{eq:inpID}
\begin{split}
    (\tilde{\bm{\sigma}}^2_n)_c&=\frac{1}{HW}\sum_{i,j}^{H,W}\left[(\gamma_c\frac{x_{ncij}-(\bm{\mu}_{\BN})_{c}}{(\bm{\sigma}_{\BN})_c}+\beta_c)-(\gamma_c\frac{(\bm{\mu}_{\IN})_{nc}-(\bm{\mu}_{\BN})_{c}}{(\bm{\sigma}_{\BN})_{c}}+\beta_c)\right]^2\\
    &=\frac{\gamma_c^2}{(\bm{\sigma}^2_{\BN})_{c}}\frac{1}{HW}\sum_{i,j}^{H,W}(x_{ncij}- (\bm{\mu}_{\IN})_{nc})^2\\
    &=\frac{\gamma_c^2(\bm{\sigma}^2_{\IN})_{nc}}{(\bm{\sigma}^2_{\BN})_{c}}
\end{split}
\end{equation}
Rewritting Eqn.(\ref{eq:inpID}) into the vector form gives us $\tilde{\bm{\sigma}}^2_{n}=\diag(\bm{\gamma}\bm{\gamma}\tran)\odot\frac{(\bm{\sigma}^2_{\IN})_{n}}{\bm{\sigma}^2_{\BN}}$, where $\diag (\bm{\gamma}\bm{\gamma}\tran) \in \R^{C\times 1}$ extracts the diagonal of the given matrix. At last,  the output of BN is $\tilde{x}_{ncij}=\gamma_c\bar{x}_{ncij}+\beta_c$, then the entry in c-th row and d-th column of covariance matrix $\Sigma$ of $\tilde{x}$ is calculated as follows:
\begin{equation}
    \Sigma_{cd}=\frac{1}{NHW}\sum^{N,H,W}_{n,i,j}(\gamma_c\bar{x}_{ncij})(\gamma_d\bar{x}_{ndij})=\gamma_c\gamma_d\rho_{cd}
\end{equation}
where $\rho_{cd}$ is the element in c-th row and j-th column of correlation matrix of $\bar{x}$. Thus, we can write $\Sigma$ into the vector form: $\bm{\Sigma}=\bm{\gamma}\bm{\gamma}\tran \odot \frac{1}{M}\bar{\bm{x}}\bar{\bm{x}}\tran$ if we reshape $\tilde{\bm{x}}$ to $\tilde{\bm{x}}\in\R^{C\times M}$ and $M=N\cdot H\cdot W$.

\subsection{Architecture of IR branch}\label{sec:subnetwork}
We denote the subnetwork in IR branch as $\tilde{f}$. Note that the activation of $\tilde{f}$ is the Sigmoid function, we formulate $\tilde{f}$ following \cite{hu2018squeeze},
\begin{eqnarray}
\tilde{f}(\bm{\sigma}_n^2)=\mathrm{Sigmoid}(\bm{W}_2\delta_1(\mathrm{LN}(\bm{W}_1\bm{\sigma}^2_n)))\label{eq:reparam2}
\end{eqnarray} 
where $\delta_1$ are ReLU activation function, $\bm{W}_1\in\R^{\frac{C}{r}\times C}$ and  $\bm{W}_2\in\R^{C\times \frac{C}{r}}$ and $r$ is reduction ratio ($r=4$ in our experiments), $\tilde{f}(\bm{\sigma}_n^2)\in (0,1)^C$ is treated as a gating mechanism in order to control the strength of the inverse square root of variance for each channel.  We see that $\tilde{f}$ is expressed by a bottleneck architecture that is able to model channel dependencies and limit model complexity. Layer normalization (LN) is used inside the bottleneck transform (before ReLU) to ease optimization. It is seen from Eqn.(\ref{eq:reparam1}) that $s^{-\frac{1}{2}}$ represents the quantity of inverse square root of variance and  $\tilde{f}(\bm{\sigma}_n^2)$ regulates the extend of instance reweighting. $\tilde{f}$ maps the instance variance to a set of channel weights. In this sense, the IR branch intrinsically introduces channel dependencies conditioned on each input.

\setcounter{prop}{0}
\section{Proof of proposition \ref{prop:equgmma}}\label{sec:appendixB}
\begin{prop}
Let $\bm{\Sigma}$ be covariance matrix of feature maps after batch normalization. Then, (1) assume that $\bm{\Sigma}_k=\bm{\Sigma}^{-\frac{1}{2}},\,\forall k=1,2,3,\cdots,T$, we have $|\hat{\gamma_c}|> |\gamma_c|, \, \forall c\in [C]$. (2) Let $\tilde{\bm{x}}_{nij}=\Diag(\bm{\gamma})\bar{\bm{x}}_{nij}+\bm{\beta}$, assume $\bm{\Sigma}$ is full-rank, then $ \left\|\bm{\Sigma}^{-\frac{1}{2}}\tilde{\bm{x}}_{nij}\right\|_2>\left\|\tilde{\bm{x}}_{nij}\right\|_2$
\vspace{-3pt}
\end{prop}

Proof. (1) Since $\bm{\Sigma}_k=\bm{\Sigma}^{-\frac{1}{2}},\,\forall k=1,2,\cdots,T$, we have $\bm{\Sigma}_k\bm{\gamma}=\frac{1}{2}\bm{\Sigma}_{k-1}(3\bm{I}-\bm{\Sigma}_{k-1}^2\bm{\Sigma})\bm{\gamma}=\bm{\Sigma}_{k-1}\bm{\gamma}$. Therefore, we only need to show $\left\|\hat{\bm{\gamma}}\right\|_1=\left\|\bm{\Sigma}_T\bm{\gamma}\right\|_1=\cdots=\left\|\bm{\Sigma}_1\bm{\gamma}\right\|_1> \left\|\bm{\gamma}\right\|_1$. Now, we show that for $k=1$ we have $\left\|\frac{1}{2}(3\bm{I}-\bm{\Sigma})\bm{\gamma}\right\|_1> \left\|\bm{\gamma}\right\|_1$. From Eqn.(\ref{eq:batch-co}), we know that $\bm{\Sigma}=\frac{\bm{\gamma}\bm{\gamma}\tran}{\left\|\bm{\gamma}\right\|_2^2} \odot\bm{\rho}$ where $\bm{\rho}$ is the correlation matrix of $\tilde{\bm{x}}$ and $-1\leq\rho_{ij}\leq 1,\,\forall i,j\in[C]$. Then, we have
\begin{equation}\label{eq:gamma1}
\begin{split}
    \frac{1}{2}(3\bm{I}-\bm{\Sigma})\bm{\gamma}&=\frac{1}{2}(3\bm{I}-\frac{\bm{\gamma}\bm{\gamma}\tran}{\left\|\bm{\gamma}\right\|_2^2} \odot\bm{\rho})\bm{\gamma}\\
    &=\frac{1}{2}(3\bm{\gamma}-(\frac{\bm{\gamma}\bm{\gamma}\tran}{\left\|\bm{\gamma}\right\|_2^2}\odot\bm{\rho})\bm{\gamma})\\
    &=\frac{1}{2}(3\bm{\gamma}-\frac{1}{\left\|\bm{\gamma}\right\|_2^2}\left[\sum_{j}^C\gamma_1\gamma_j\rho_{1j}\gamma_j,\sum_{j}^C\gamma_2\gamma_j\rho_{2j}\gamma_j,\cdots,\sum_{j}^C\gamma_C\gamma_j\rho_{Cj}\gamma_j\right]^{\mathrm{T}})\\
     &=\frac{1}{2}(3\bm{\gamma}-\frac{1}{\left\|\bm{\gamma}\right\|_2^2}\left[\sum_{j}^C\gamma_1\gamma_j\rho_{1j}\gamma_j,\sum_{j}^C\gamma_2\gamma_j\rho_{2j}\gamma_j,\cdots,\sum_{j}^C\gamma_C\gamma_j\rho_{Cj}\gamma_j\right]^{\mathrm{T}})\\
     &=\frac{1}{2}\left[(3-\sum_{j}^C\frac{\gamma_j^2\rho_{1j}}{\left\|\bm{\gamma}\right\|_2^2})\gamma_1,(3-\sum_{j}^C\frac{\gamma_j^2\rho_{2j}}{\left\|\bm{\gamma}\right\|_2^2})\gamma_2,\cdots,(3-\sum_{j}^C\frac{\gamma_j^2\rho_{Cj}}{\left\|\bm{\gamma}\right\|_2^2})\gamma_C\right]^{\mathrm{T}}
\end{split}
\end{equation}
Note that $|3-\sum_{j}^C\frac{\gamma_j^2\rho_{ij}}{\left\|\bm{\gamma}\right\|_2^2}|\geq 3-|\sum_{j}^C\frac{\gamma_j^2\rho_{ij}}{\left\|\bm{\gamma}\right\|_2^2}|\geq 3- \sum_{j}^C\frac{\gamma_j^2}{\left\|\bm{\gamma}\right\|_2^2}=2$, where the last equality holds iff $\rho_{ij}=1,\,\forall i,j\in [C]$. However, this is not the case in practice. Hence, for all $c\in[C]$ we have 
\begin{equation}
    \left|\left[\frac{1}{2}(3\bm{I}-\bm{\Sigma})\bm{\gamma}\right]_c\right|=\left|\frac{1}{2}(3-\sum_{j}^C\frac{\gamma_j^2\rho_{cj}}{\left\|\bm{\gamma}\right\|_2^2})\gamma_c\right|> |\gamma_c|
\end{equation}
  Note that if other normalization methods such as IN and LN are used, the conclusion in Proposition \ref{prop:equgmma}  can be still drawn as long as the condition $-1\leq\rho_{ij}\leq 1,\,\forall i,j\in[C]$ is satisfied.
 
 (2) We first show that $\lambda_i\in (0,1), \forall i\in [C]$ where $\lambda_i$ is the $i$-th largest eigenvalues of $\bm{\Sigma}$. Since $\bm{\Sigma}$ is the covariance matrix and has full rank, we have $\lambda_i>0$. Moreover, $\sum_{i=1}^{C}\lambda_i=\mathrm{tr}(\bm{\Sigma})=1$. Hence, we obtain that $\lambda_i\in (0,1)$. Based on this fact, the inequality below can be derived,
 \begin{equation}
     \left\|\bm{\Sigma}^{-\frac{1}{2}}\tilde{\bm{x}}_{nij}\right\|_2^2=\tilde{\bm{x}}_{nij}\tran\bm{\Sigma}^{-1}\tilde{\bm{x}}_{nij}> \frac{1}{\lambda_1}\tilde{\bm{x}}_{nij}\tran \tilde{\bm{x}}_{nij}>\tilde{\bm{x}}_{nij}\tran \tilde{\bm{x}}_{nij}=\left\|\tilde{\bm{x}}_{nij}\right\|_2^2
 \end{equation}
 Hence, $ \left\|\bm{\Sigma}^{-\frac{1}{2}}\tilde{\bm{x}}_{nij}\right\|_2>\left\|\tilde{\bm{x}}_{nij}\right\|_2$. Here completes the proof.

\section{Connection between CE block and Nash Equilibrium}\label{sec:appendixC}
We first introduce the definition of Gaussian interference game in context of CNN and then build the connection between a CE block and Nash Equilibrium. For clarity of notation, we omit the subscript $n$ for a concrete sample.

%
%
We suppose that each channel can transmit a power vector $\bm{p}_c=(p_{c11},\cdots,p_{cHW})$ where $p_{cij}$ denotes the transmit power to the neuron at location $(i,j)$ in the $c$-th channel. Since normalization layer is often followed by a ReLU activation, we restrict $p_{cij}\geq 0$.  
In game theory, all channels maximizes sum of maximum information rate of all neurons. We consider dependencies among channels, then channels are thought to play a Gaussian interference game, which can be described by the following maximization problem, for the $c$-th channel,
\begin{equation}\label{payoffc}
\begin{split}
&\max \,\, \mathcal{C}_c(\bm{p}_1,\bm{p}_2,\cdots,\bm{p}_C)=\sum_{i,j=1}^{h,W}\ln \left(1+\frac{g_{cc}p_{cij}}{\sum_{d\neq c} g_{cd}p_{dij}+\sigma_c/h_{cij}}\right)\\
&s.t.\quad  \left\{\begin{array}{lc}
\sum_{i,j=1}^{H,W}p_{cij}= P_c,\\
p_{cij}\geq 0,\,& \forall i\in [H],j\in [W]
\end{array}
\right.
\end{split}
\end{equation}
where $g_{cd}$ represents dependencies between the $c$-th channel and $d$-th channel, and $\mathcal{C}_c$ is the sum of maximum information rate with respect to to the $c$-th channel given transit power distributions $\bm{p}_1,\bm{p}_2,\cdots,\bm{p}_C$. We also term it pay-off of the $c$-th channel. In game theory, $C$ channels and solution space of $\{p_{cij}\}_{c,i,j=1}^{C,H,W}$ together with pay-off vector $\bm{\mathcal{C}}=(\mathcal{C}_1,\mathcal{C}_2,\cdots,\mathcal{C}_C)$ form a Gaussian interference game $\mathbb{G}$.  Different from basic settings in $\mathbb{G}$, here we do not restrict dependencies $g_{cd}$ to $(0,1)$. It is known that $\mathbb{G}$ has a unique Nash Equilibrium point whose definition is given as below,
\begin{Def}\label{nashdef}
An $C$-tuple of strategies $(\bm{p}_1,\bm{p}_2,\cdots,\bm{p}_C)$ for channels $1,2,\cdots,C$ respectively is called a Nash equilibrium iff for all $c$ and for all $\bm{p}$ ($\bm{p}$ a strategy for channel $c$)
\begin{equation}\label{nasheqn}
    \mathcal{C}_c(\bm{p}_1,\cdots,\bm{p}_{c-1},\bm{p},\bm{p}_{c+1},\cdots,\bm{p}_C)\leq \mathcal{C}_c(\bm{p}_1,\bm{p}_2,\cdots,\bm{p}_C)
\end{equation}
\end{Def}
i.e., given that all other channels $d\neq c$ use strategies $\bm{p}_d$, channel $c$ best response is $\bm{p}_c$. Since $\mathcal{C}_1,\mathcal{C}_2,\cdots,\mathcal{C}_C$ are concave in $\bm{p}_1,\bm{p}_2,\cdots,\bm{p}_C$ respectively, KKT conditions imply the following theorem.
\begin{theorem}\label{nashtheo}
Given pay-off in Eqn.(\ref{payoffc}), $(\bm{p}_1^*,\cdots,\bm{p}_C^*)$ is a Nash equilibrium point if and only if there exist $\bm{v}_0=(v_0^1,\cdots,v_0^C)$ (Lagrange multiplier) such that for all $i\in [H]$ and $j\in [W]$,
\begin{equation}\label{nashpoint}
    \frac{g_{cc}}{\sum_d g_{cd}p_{dij}^*+\sigma_c/h_{cij}}\left\{\begin{array}{lc}
=v_0^c\,\mathrm{for}\,p_{cij}^*>0\\
\leq v_0^c\,\mathrm{for}\, p_{cij}^*=0\\
\end{array}\right.
\end{equation}
\end{theorem}

Proof. The Lagrangian corresponding to minimization of $-C_c$ subject to the equality constraint and non-negative constraints on $p_{cij}$ is given by
\begin{equation}
    L_c=-\sum_{i,j=1}^{h,W}\ln \left(1+\frac{g_{cc}p_{cij}}{\sum_{d\neq c} g_{cd}p_{dij}+\sigma_c/h_{cij}}\right)+v_0^c(\sum_{i,j=1}^{H,W}p_{cij}- P_c)+\sum_{i,j=1}^{H,W}v_1^{cij}(-p_{cij}).
\end{equation}
Differentiating the Lagrangian with respect to $p_{cij}$ and equating the derivative to zero, we obtain
\begin{equation}
\frac{g_cc}{\sum_dg_{cd}p_{cij}+\sigma_c/h_{cij}}+v_1^{cij}=v_0^c
\end{equation}
Now, using the complementary slackness condition $v_1^{cij}p_{cij}=0$ and $v_1^{cij}\geq 0$, we obtain condition (\ref{nashpoint}). This completes the proof.

By Theorem \ref{nashtheo}, the unique Nash Equilibrium point can be explicitly written as follows when $p^*_{cij}>0$,
\begin{equation}\label{eq:nash-solution}
\bm{p}^*_{ij}=\bm{G}^{-1}\left(\Diag(\bm{v}_0)^{-1}\diag(\bm{G})-\Diag(\bm{h}_{ij})^{-1}\bm{\sigma} \right)
\end{equation}
where $\bm{p}^*_{ij}, \bm{h}_{ij}, \bm{\sigma} \in \R^{C\times 1}$ and $\bm{v}_0\in \R^{C\times 1}$ are Lagrangian multipliers corresponding to equality constraints. Note that a approximation can be made  using Taylor expansion as follow: $-\frac{\sigma_c}{h_{cij}}=\sigma_c(2+ h_{cij}+\mathcal{O}((1+h_{cij})^2))$. Thus, a linear proxy to Eqn.(\ref{eq:nash-solution}) can be written as 
\begin{equation}\label{eq:nash-solution-linear}
\bm{p}^*_{ij}=\bm{G}^{-1}\left(\Diag(\bm{\sigma})\bar{\bm{h}}_{ij}+\Diag(\bm{v}_0)^{-1}\diag(\bm{G})+(2+\bm{\delta})\bm{\sigma}\right)
\end{equation}
Let $\bm{G}=[\bm{D}_n]^\frac{1}{2}, \bm{h}_{ij}=\bar{\bm{x}}_{ij}, \bm{\gamma}=\bm{\sigma}$ and $\bm{\beta}=\Diag(\bm{v}_0)^{-1}\diag(\bm{G})+(2+\bm{\delta})\bm{\sigma}$, Eqn.(\ref{eq:nash-solution-linear}) can surprisingly match CE unit in Eqn.(\ref{eq:CE}), implying that the proposed CE block indeed encourages all the channels to contribute to the layer output. In Gaussian interference game, $\bm{\sigma}$ is known and $v_0$ can be determined when budget $P_c$'s are given. However, $\bm{\gamma}$ and $\bm{\beta}$ are learned by SGD in deep neural networks. Note that the Nash Equilibrium solution can be derived for every single sample, implying that the decorrelation operation should be performed conditioned on each instance sample. This is consistent with our design of the CE block.

\section{Network Architecture}\label{sec:netmobile}
\begin{figure}
    \centering
    \includegraphics[width=0.5\textwidth]{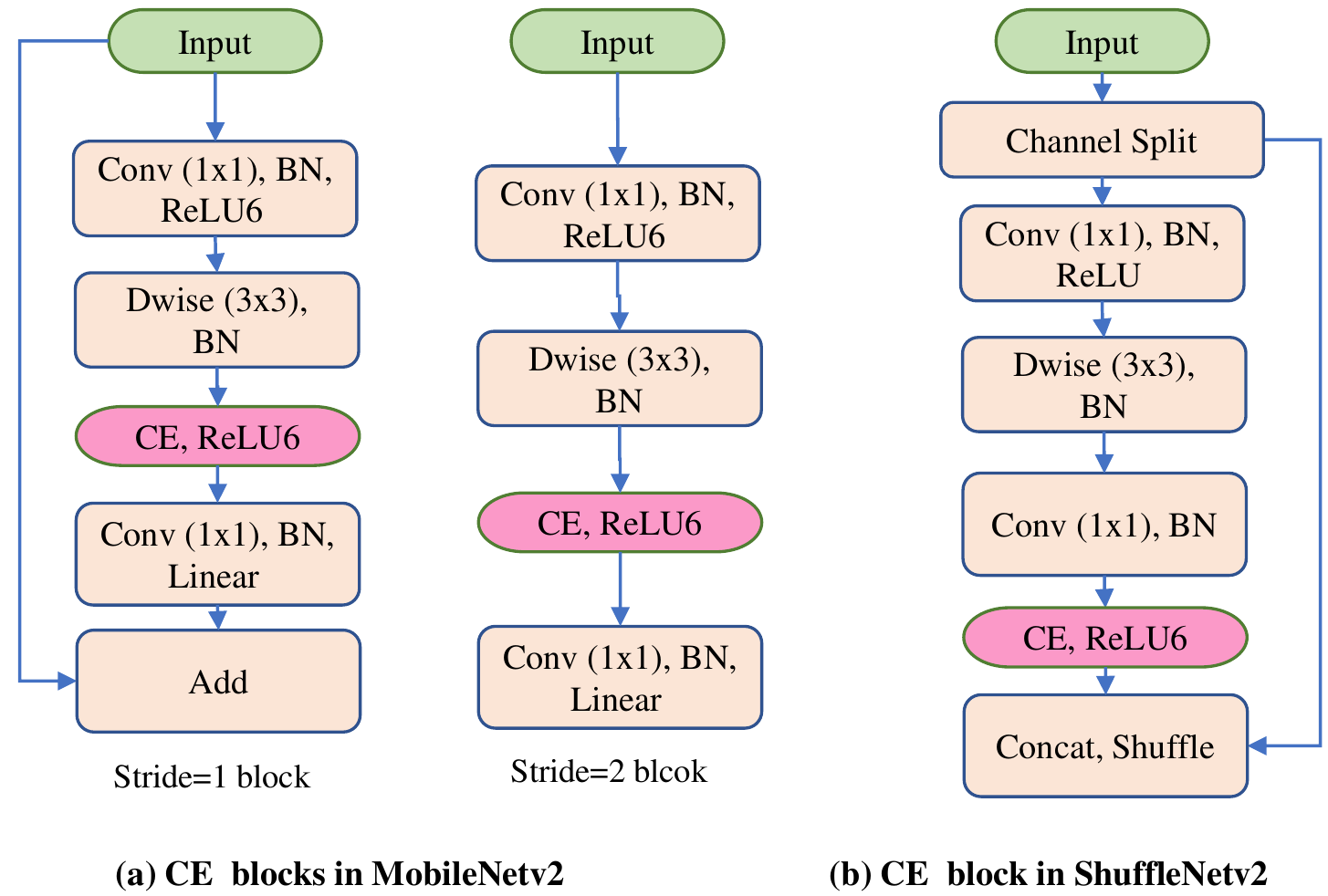}
    \caption{CE blocks in MobileNetv2 (a) and ShuffleNetv2 (b). `Add' denotes broadcast element-wise addition. `Concat' indicates the concatenation of channels. `Dwise' represents the depthwise convolution.}
    \label{fig:netmobileshuffle}
\end{figure}

\textbf{CE-MobileNetv2 and CE-ShuffleNetv2.} As for CE-MobileNetv2, since the last `$1\times 1$' convolution layer in the bottleneck is not followed by a Rectified unit, we insert CE in the `$3\times 3$' convolution layer which also has the largest number of channels in the bottleneck, as shown in Fig.\ref{fig:netmobileshuffle}(a). Following similar strategies, CE is further integrated into ShuffleNetv2 to construct CE-ShuffleNetv2, as shown in Fig.\ref{fig:netmobileshuffle}(b). 
\begin{table}
\vspace{-0.2cm}
\centering
\small
\begin{tabular}{c| c c| c c| c c|c c|c c}
\hline
    Backbone &\multicolumn{6}{c|}{ResNet50}&\multicolumn{4}{c}{ResNet18}\\
    \hline
         Block&\multicolumn{2}{c|}{GN+ReLU}&\multicolumn{2}{c|}{IN+ReLU}&\multicolumn{2}{c|}{LN+ReLU}&\multicolumn{2}{c|}{BN+ReLU}&\multicolumn{2}{c}{BN+ELU}\\
    
    \hline
     Acc &top-1 &top-5 &top-1 &top-5 &top-1 &top-5 &top-1 &top-5 &top-1 &top-5 \\
    Baseline  &75.6 &92.8 &74.2 &91.9 &71.6 &89.9 &70.5 &89.4 &68.1 &87.6\\
    Baseline+CE &\textbf{76.2} &\textbf{92.9} &\textbf{76.0}&\textbf{92.7}&\textbf{73.3}&\textbf{91.3}&\textbf{71.9}&\textbf{90.2}&\textbf{68.7}&\textbf{88.5}\\
    Increase &+0.6&+0.1&+1.8&+0.8&+1.7&+1.4 &+1.4&+0.8&+0.6&+0.9 \\
    \hline
    \end{tabular}
    \centering
    \caption{CE improves top-1 and top-5 accuracy of various normalization methods and rectified units on ImageNet with ResNet50 or ResNet18 as backbones. }
    \label{tab:other-normalizer}
\hfill
\centering
\vspace{-0.4cm}
\end{table}

\textbf{Moving average in inference.}
Unlike previous methods in manual architecture design that do not depend on batch estimated statistics, the proposed CE block requires computing the inverse square root of a batch covariance matrix $\bm{\Sigma}$ and a global variance scale $s$ in Eqn.(\ref{eq:reparam1}) in each training step. To make the output depend only on the input, deterministically in inference, we use the moving average to calculate the population estimate of $\hat{\bm{\Sigma}}^{-\frac{1}{2}}$ and $\hat{s}^{-\frac{1}{2}}$  by following the below updating rules:
\begin{equation}\label{eq:movinginfernce}
    \hat{\bm{\Sigma}}^{-\frac{1}{2}}=(1-m)\hat{\bm{\Sigma}}^{-\frac{1}{2}}+m\bm{\Sigma}^{-\frac{1}{2}},\,\,\,\hat{s}^{-\frac{1}{2}}=(1-m)\hat{s}^{-\frac{1}{2}}+m\cdot s^{-\frac{1}{2}}
\end{equation}
where $s$ and $\bm{\Sigma}$ are the variance scale and covariance calculated within each mini-batch during training, and $m$ denotes the momentum of moving average. It is worth noting that since $\hat{\bm{\Sigma}}^{-\frac{1}{2}}$ is fixed during inference, the BD branch does not introduce extra costs in memory or computation except for a simple linear transformation ( $\hat{\bm{\Sigma}}^{-\frac{1}{2}}\tilde{\bm{x}}_{nij}$).

\textbf{Model and computational complexity}.
The main computation of our CE includes calculating the covariance and inverse square root of it in the BD branch and computing two FC layers in the IR branch. We see that there is a lot of space to reduce computational cost of CE. For BD branch, given an internal feature $\bm{x}\in \R^{N\times C\times H\times W}$, the cost of calculating a covariance matrix is $2NHWC^2$, which is comparable to the cost of convolution operation. A pooling operation can be employed to downsample featuremap for too large $H$ and $W$. In this way, the complexity can be reduced to $2NHWC^2/k^2+CHW$ where $k$ is kernel size of the window of pooling. Further, we can use group-wise whitening to improve efficiency, reducing the cost of computing $\bm{\Sigma}^{-\frac{1}{2}}$ from $TC^3$ to $TCg^2$ ($g$ is group size). For IR branch, we focus on the additional parameters introduced by two FC layers. In fact, the reduction ratio $r$ can be appropriately chosen to balance model complexity and representational power. Besides, the majority of these parameters come from the final block of the network. For example, a single IR in the final block of ResNet-50 has $2*2048^2/r$ parameters. In practice, the CE blocks in the final stages of networks are removed to reduce additional parameters. We provide the measurement of computational burden and Flops in Table \ref{tab:acc-resent}.

\section{Ablative Experiments}\label{sec:appendixD}
\begin{table}
    \centering
    \small
\begin{tabular}{c c c c c}
    \hline
         ResNet50&Baseline &+BD&+IR&+CE\\
    \hline
         top-1 &76.6&77.0 &77.3 &\textbf{78.3 (+1.7)}\\
    \hline
    \end{tabular}
    \vspace{-0.1cm}
    \caption{Results of BD, IR and CE on Imagenet with ResNet-50 as the basic structure. The top-1 accuracy increase (1.7) of CE-ResNet is higher than combined top-1 accuracy increase (1.1) of BD-ResNet and IR-ResNet, indicating the effects of BD and IR branch is complementary.}
    \label{tab:BCDAVICE}
    \vspace{-0.1cm}
\end{table}
\begin{table}
\vspace{-0.2cm}
\centering
\small
\begin{tabular}{c c c c c c c}
\hline
    Backbone &\multicolumn{3}{c}{ResNet50}&\multicolumn{3}{c}{ResNet18}\\
    \hline
         Method&{IterNorm}&{SW}&{CE}&{IterNorm}&{SW}&{CE}\\ 
    \hline
     Top-1 &77.1 &77.9 &\textbf{78.3} &71.1 &71.6 &\textbf{71.9}\\
    \hline
    \end{tabular}
    \centering
    \caption{Comparison between the proposed CE and other normalization method using decorrelation on ImageNet dataset. CE achieves higher top-1 accuracy on both ResBet50 and ResNet18.}
    \label{tab:comparewithsw}
\hfill
\centering
\vspace{-0.4cm}
\end{table}
\begin{table}[]
    \centering
    \begin{tabular}{c c}
    \hline
    	& Top-1 acc\\
    \hline
         CE2-ResNet50& 77.9 \\
         CE3-ResNet50&78.3 \\ 
    \hline
    \end{tabular}
    \caption{We add CE after the second (CE2-ResNet50) and third (CE3-ResNet50) batch normalization layer in each residual block. The channel of the third batch normalization is 4 times than that of the second one but the top-1 accuracy of CE3-ResNet50 outperforms CE2-ResNet50 by 0.4, which indicates CE benefits from larger number of channels.}
    \label{tab:integrationstra}
\end{table}

\textbf{CE improves various normalization methods and rectified units.}
In addition to BN, CE is also effective for other normalization technologies, as inhibited channel emerges in many well-known normalizers as shown in Fig.\ref{fig1:sparse-top1first}.
To prove this, we conduct experiments using ResNet-50 under different normalizers including, group normalization (GN), instance normalization (IN), and layer normalization (LN). 
For these experiments, we stack CE block after the above normalizers to see whether CE helps other normalization methods. Table \ref{tab:other-normalizer} confirms that CE generalize well over different normalization technologies, improving their generalization on testing samples by 0.6-1.8 top-1 accuracy. On the other hand, CE is also superior to many rectified units such as ELU \citep{clevert2015fast}

\begin{figure}
    \centering
\includegraphics[width=0.8\textwidth]{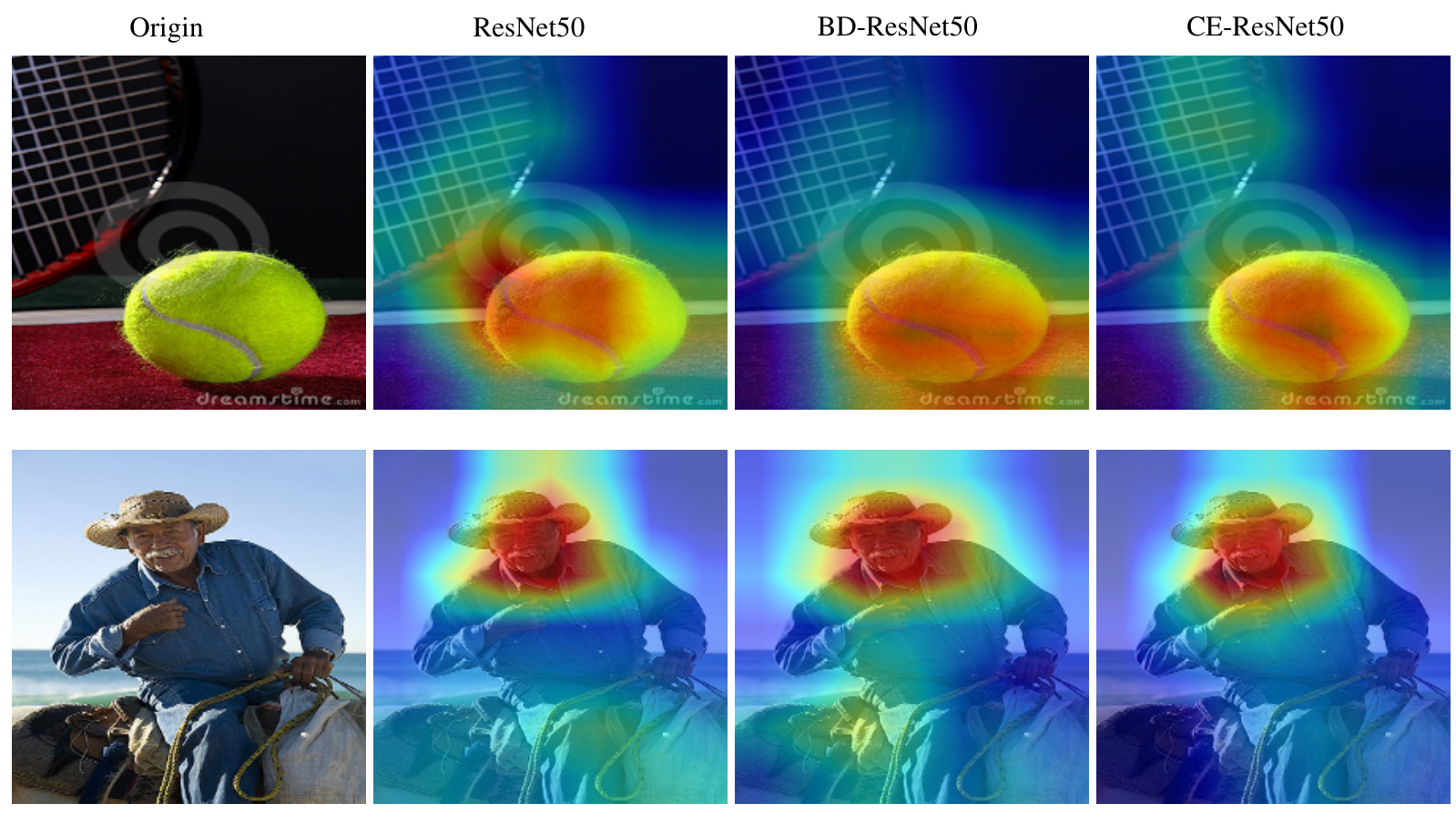}
\caption{The Grad-cam~\cite{selvaraju2017grad} visualization results from the final convolutional layer for plain ResNet50, SE-ResNet50, and CE-ResNet50. We see that heat maps extracted from CAM for CE-ResNet50 have more coverage on the object region and less coverage on the background region than that of BD-ResNet50, implying that IR can help CE learn preciser features.}
\label{fig:inputIDbranch}
\end{figure}

\textbf{IR helps CE learn preciser feature representation.} 
%
The IR branch adjusts the correlations among feature channels for each instance sample, it is expected to make the network respond to different inputs in a highly class-specific manner. In this way, it helps CE learn preciser feature representation. To verify this, we employ an off-the-shelf tool to visualize the class activation map (CAM) \citep{selvaraju2017grad}. We use ResNet50, BD-ResNet50, and CE-ResNet50 trained on ImageNet for comparison. As shown in Fig.\ref{fig:inputIDbranch}, the heat maps extracted from CAM for CE-ResNet50 have more coverage on the object region and less coverage on the background region. It shows that the IR branch helps CE learn preciser information from the images.


\textbf{BD and IR are complementary}.
Here, we verify that BD and IR are complementary to each other. We train plain ResNet50, BD-ResNet50, IR-ResNet50, and CE-ResNet50 for comparison. The top-1 accuracy is reported in Table \ref{tab:BCDAVICE}. It is observed that the BD-ResNet50 and IR-ResNet50 are 0.4 and 0.7 higher than the plain ResNet-50 respectively. However, when they are combined, the top-1 accuracy improves by 1.7, higher than combined accuracy increase (1.1), which demonstrates that they benefit from each other. 

\textbf{Integration strategy of CE block.} We put CE in different position of a bottleneck in ResNet50, which consists of three "Conv-BN-ReLU" basic blocks. The channel of the third block is 4 times than that of the second one. We compare the performance of CE-ResNet50 by putting CE in the second block (CE2-ResNet50) or the third block (CE3-ResNet50). As shown in Table  \ref{tab:integrationstra}, the top-1 accuracy of CE3-ResNet-50 outperforms CE2-ResNet50 by 0.4, which indicates that our CE block benefits from larger number of channels.

\textbf{Comparison with normalization methods using decorrelation.} Many normalization approaches also use decorrelation operation such as switchable whitening (SW) \citep{pan2019switchable} and IterNorm \citep{huang2019iterative} to stabilize the course of training . However, all of them are applied after convolution layer. Thus, the inhibited channels still emerge due to the use of affine transformation (\ie $\bm{\gamma}$ and $\bm{\beta}$. Instead, our proposed CE decorrelates features after normalization layer conditioned on each instance, which has been proved to be able to prevent inhibited channels. Here we show that CE obtains a gain of performance on ImageNet with ResNet18 and ResNet50 compared with SW and IterNorm. The results are repoted in Table \ref{tab:comparewithsw}.


\subsection{More discussion about CE}
Many methods have been proposed to improve normalizers such as switchable normalization (SN) \citep{luo2018differentiable} and ReLU activation such as  exponential linear unit (ELU) \citep{clevert2015fast} and leaky ReLU (LReLU) \citep{maas2013rectifier}. The ablation approach in \cite{morcos2018importance} is used to see whether and how these methods encourage channels to contribute equally to learned feature representation. Here we call this property `channel equalization' for clarity of narration. We demonstrate the effectiveness of CE by answering the following questions. 
\begin{figure}
\begin{subfigure}{.5\textwidth}
\centering
\includegraphics[width=0.7\textwidth]{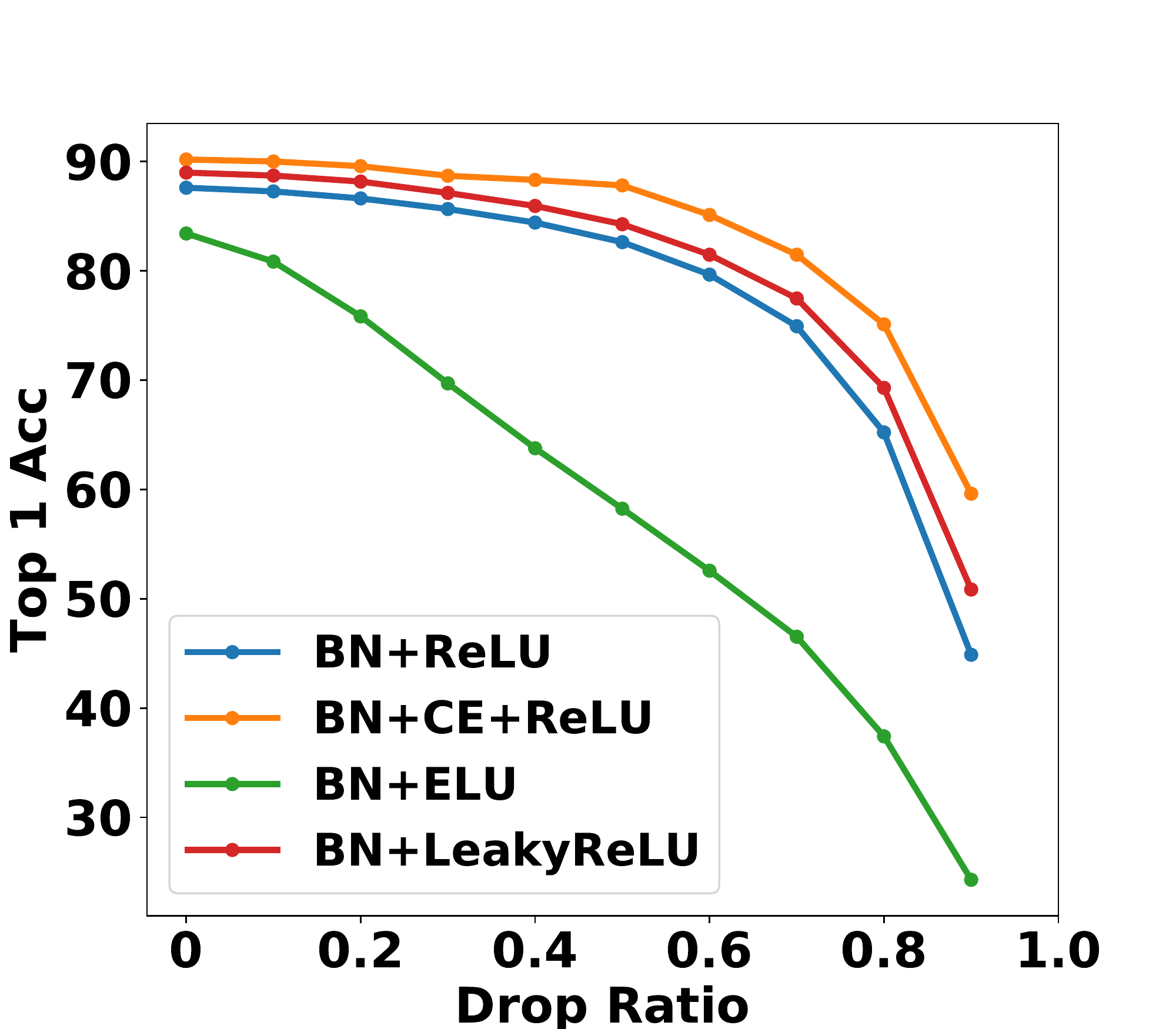}
\caption{}
\end{subfigure}
\begin{subfigure}{.5\textwidth}
\centering
\includegraphics[width=0.7\textwidth]{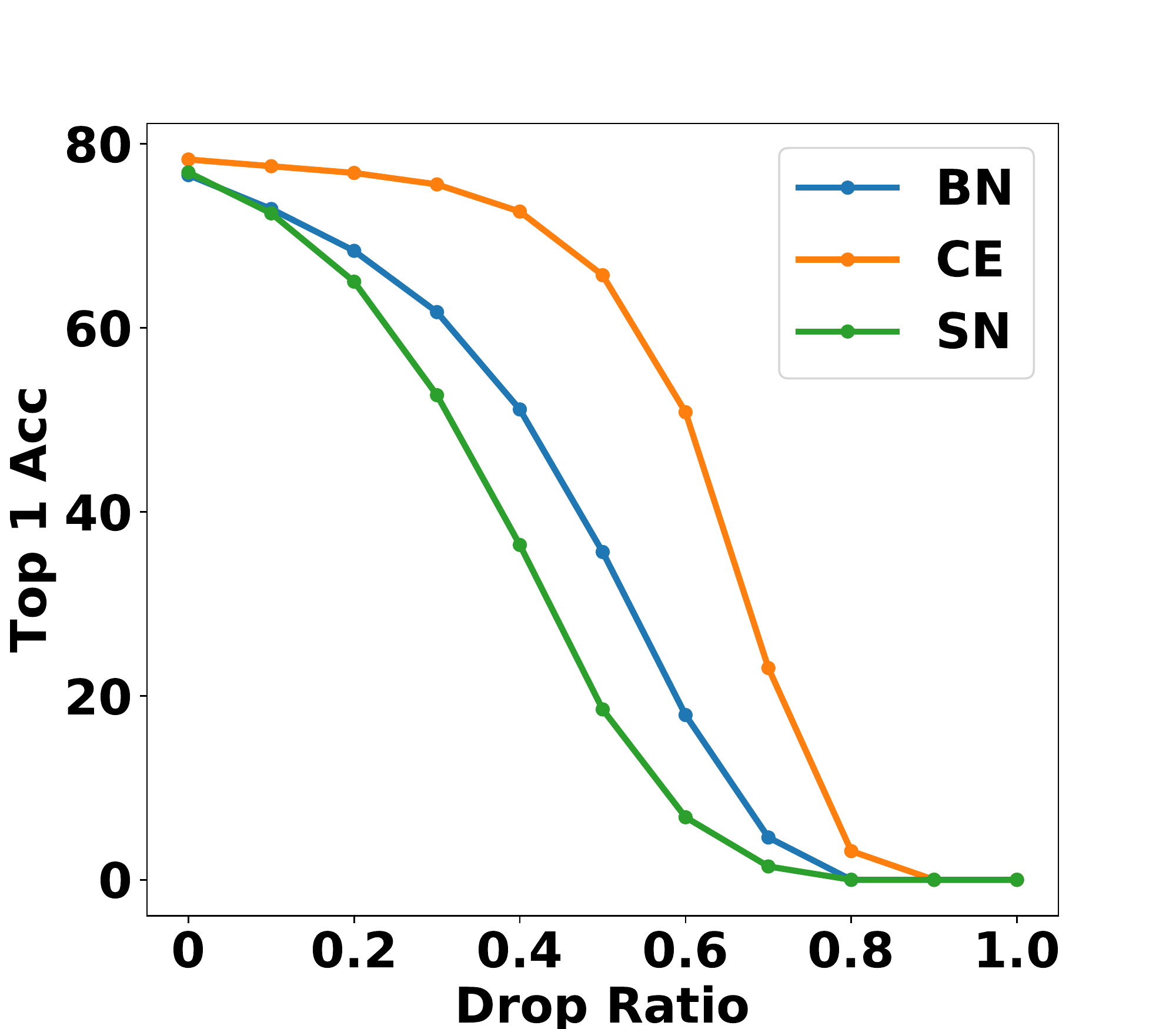}
\caption{}
\end{subfigure}
\vspace{-0.1cm}
\caption{(a) compares the cumulative ablation curves of 'BN+ReLU', 'BN+ELU', 'BN+LReLU' and 'BN+CE+ReLU' with VGGNet on CIFAR-10 dataset. We see that the Both LReLU and  CE can improve the channel equalization in 'BN+ReLU' block. (b) compares the cumulative ablation curves of 'BN+ReLU', 'SN+ReLU' and 'BN+CE+ReLU' with ResNet-50 on ImageNet dataset. The proposed CE consistently improves the channel equalization of 'BN+RelU' block. Note that 'BN+CE+ReLU' achieves the highest top-1 accuracy on both two datasets compared to its counterparts (when drop ration is $0$). }\label{fig:relu-norm-drop}
\vspace{-0.4cm}
\end{figure}

\textbf{Do other ReLU-like activation functions help channel equalization?} Two representative improvements on ReLU function, i.e. ELU \citep{clevert2015fast} and LReLU \citep{maas2013rectifier}, are employed to see whether other ReLU-like activation functions can help channel equalization.  We plot the cumulative ablation curve that depicts ablation ratio versus the top-1 accuracy on CIFAR10 dataset in Fig.\ref{fig:relu-norm-drop}(a). The baseline curve is 'BN+ReLU'. As we can see, the top-1 accuracy curve of 'BN+LReLU' drops more gently, implying that LReLU helps channel equalization. But 'ELU+ReLU' has worse cumulative ablation curve than 'BN+ReLU'. By contrast, the proposed CE block improves the recognition performance of 'BN+ReLU' (higher top-1 accuracy) and promotes channel equalization most (the most gentle cumulative ablation curve).


\textbf{Do the adaptive normalizers encourage channels?} We experiment on a representative adaptive normalization method (i.e. SN), to see whether it helps channel equalization. SN learns to select an appropriate normalizer from IN, BN and LN for each channel. The cumulative ablation curves are plotted on ImageNet dataset with ResNet-50 under blocks of 'BN+ReLU', ‘SN+ReLU' and 'BN+CE+ReLU'. As shown in Fig.\ref{fig:relu-norm-drop}(b), SN even does damage to channel equalization when it is used to replace BN. However, 'BN+CE+ReLU' shows the most gentle cumulative ablation curve, indicating the effectiveness of CE block in channel equalization. Compared with SN, ResNet-50 with CE block also achieves better top-1 accuracy (78.3 vs 76.9), showing that channel equalization is important for block design in a CNN.

\section{Experimental Setup}\label{sec:appendixE}
\textbf{ResNet Training Setting on ImageNet}.
All networks are trained using 8 GPUs with a mini-batch of $32$ per GPU. We train all the architectures from scratch for $100$ epochs using stochastic gradient descent (SGD) with momentum $0.9$ and weight decay 1e-4. The base learning rate is set to $0.1$ and is multiplied by 0.1 after $30,60$ and $90$ epochs. Besides, the covariance matrix in BD branch is calculated within each GPU. Since the computation of covariance matrix involves heavy computation when the size of feature map is large, a $2\times 2$ maximum pooling is adopted to down-sample the feature map after the first batch normalization layer. Like \citep{huang2019iterative}, we also use group-wise decorrelation with group size $16$ across the network to improve the efficiency in the BD branch. By default, the reduction ratio $r$ in IR branch is set to $4$.

\textbf{MobileNet V2 training setting on ImageNet}.
All networks are trained using 8 GPUs with a mini-batch of $32$ per GPU for 150 epochs with cosine learning rate. The base learning rate is set to 0.05 and the weight decay is 4e-5. 

\textbf{ShuffleNet V2 training setting on ImageNet}.
All networks are trained using 8 GPUs with a mini-batch of $128$ per GPU for 240 epochs with poly learning rate. The base learning rate is set to 0.5 and weight decay is 4e-5. We also adopt warmup and label smoothing tricks. 

\textbf{VGG networks training setting on CIFAR10}. We adopt CIFAR10 that
contains 60k images of 10 categories, where 50k images for
training and 10k images for test. We train VGG networks with a batch size of $256$ on a single GPU for $160$ epochs. The initial learning rate is 0.1 and is decreased by $10$ times every $60$ epochs. The inhibited channel ratios in Fig. \ref{fig1:sparse-top1first} and Fig.\ref{fig:CEeuqal}(c) is measured by the average ratio for the first six layers. For inference drop experiments in Fig.\ref{fig1:sparse-top1first}(c), we randomly drop channels of the output in the third layer with different dropout ratio. For each ratio, we run the experiment 5 times and average the top 1 accuracy.

\textbf{Mask-RCNN training setting on COCO}.
We fine-tune the ImageNet pretrained model in COCO for 24 epoch with base learning rate 0.02 and multiply it by 0.1 after 16 and 22 epochs. All the models are trained using 8 GPUs with a mini-batch of 2 images. The basic backbone structure is adopted from the ResNet50/ResNet101 trained on ImageNet.

\end{document}